\documentclass[letterpaper]{article} % DO NOT CHANGE THIS
\usepackage{aaai24}  % DO NOT CHANGE THIS
\usepackage{times}  % DO NOT CHANGE THIS
\usepackage{helvet}  % DO NOT CHANGE THIS
\usepackage{courier}  % DO NOT CHANGE THIS
\usepackage[hyphens]{url}  % DO NOT CHANGE THIS
\usepackage{graphicx} % DO NOT CHANGE THIS
\usepackage{natbib}  % DO NOT CHANGE THIS AND DO NOT ADD ANY OPTIONS TO IT
\usepackage{caption} % DO NOT CHANGE THIS AND DO NOT ADD ANY OPTIONS TO IT
\usepackage{algorithm}
\usepackage{algorithmic}
\usepackage{soul}
\usepackage{graphicx}
\usepackage{amsmath}
\usepackage{amsthm}
\usepackage{booktabs}
\usepackage{amsfonts}
\usepackage{subfigure}
\usepackage{multirow}
\usepackage{color}
\usepackage{makecell}
\usepackage{enumitem} 
\usepackage{pifont}
\usepackage{newfloat}
\usepackage{listings}
\DeclareCaptionStyle{ruled}{labelfont=normalfont,labelsep=colon,strut=off} % DO NOT CHANGE THIS
\floatstyle{ruled}
\newfloat{listing}{tb}{lst}{}
\floatname{listing}{Listing}
\title{DCLP: Neural Architecture Predictor with Curriculum Contrastive Learning }
\author{
    Shenghe Zheng, Hongzhi Wang~\thanks{Corresponding author.}, Tianyu Mu
}
\affiliations{
    Massive Data Computing Lab, Harbin Institute of Technology\\
    shenghez.zheng@gmail.com, \{wangzh, mutianyu\}@hit.edu.cn
}

\usepackage{bibentry}
\begin{document}

\maketitle

\begin{abstract}
Neural predictors have shown great potential in the evaluation process of neural architecture search (NAS). However, current predictor-based approaches overlook the fact that training a predictor necessitates a considerable number of trained neural networks as the labeled training set, which is costly to obtain. Therefore, the critical issue in utilizing predictors for NAS is to train a high-performance predictor using as few trained neural networks as possible. Although some methods attempt to address this problem through unsupervised learning, they often result in inaccurate predictions. We argue that the unsupervised tasks intended for the common graph data are too challenging for neural networks, causing unsupervised training to be susceptible to performance crashes in NAS. To address this issue, we propose a \textbf{C}urricu\textbf{L}um-guided \textbf{C}ontrastive \textbf{L}earning framework for neural \textbf{P}redictor (DCLP). Our method simplifies the contrastive task by designing a novel curriculum to enhance the stability of unlabeled training data distribution during contrastive training. Specifically, we propose a scheduler that ranks the training data according to the contrastive difficulty of each data and then inputs them to the contrastive learner in order. This approach concentrates the training data distribution and makes contrastive training more efficient. By using our method, the contrastive learner incrementally learns feature representations via unsupervised data on a smooth learning curve, avoiding performance crashes that may occur with excessively variable training data distributions. We experimentally demonstrate that DCLP has high accuracy and efficiency compared with existing predictors, and shows promising potential to discover superior architectures in various search spaces when combined with search strategies. Our code is available at: https://github.com/Zhengsh123/DCLP.
\end{abstract}

\section{Introduction}
To reduce the design cost of Deep Neural Networks (DNNs), NAS has emerged as a promising technique to automatically design neural networks for specific tasks~\cite{zoph2017neural,qin2022nas}. NAS has been widely adopted for various tasks such as image classification~\cite{li2023zico} and semantic segmentation~\cite{zhang2021dcnas}. In typical NAS, searched architectures are evaluated to guide search~\cite{he2021automl}. However, the estimation is often expensive as it usually requires training DNNs from scratch to assess their performance. For instance, NASNet~\cite{zoph2018learning} needs around 40K GPU hours for estimation on CIFAR-10, making it unaffordable for many applications. Hence, predictor-based NAS~\cite{ning2020generic} has gained popularity due to its close-to-zero estimation costs and satisfactory results.

For predictor-based NAS, a key challenge is the requirement for trained neural networks to train the predictor. Training a neural network to converge may take several days, and then obtaining a substantial amount of training data is costly. Two solutions have been proposed to overcome this challenge. The first is to optimize the predictor to better utilize limited labeled data~\cite{chen2021not}, while the second is to optimize the training method by incorporating unsupervised learning to utilize unlabeled data obtained at a low cost~\cite{tang2020semi}. However, optimizing the predictor disregards the upper limit to the predictive ability learned from limited labeled data. Then, they still require a great number of training data ~\cite{ning2020generic}. On the other hand, although the introduction of unsupervised learning reduces the need for labeled data, it does not enable predictors to achieve desired performance~\cite{ijcai2022p432}.

Unsupervised neural predictors fail because the pretext tasks designed for regular graph data are too complex for neural networks~\cite{ijcai2022p432}. For example, contrastive learning, a popular unsupervised approach, requires the model to differentiate data similarity and provide similar embeddings for similar data~\cite{he2020momentum}. However, determining the similarity between neural networks is challenging even for experienced experts. Moreover, randomly selecting samples in vanilla contrastive learning creates excessive data distribution differences during training, which further complicates the learning process. Therefore, employing unsupervised learning meant for other data to neural predictors results in poor performance~\cite{tang2020semi}.
\begin{figure*}  
\centering  
\includegraphics[height=5.8cm,width=16.9cm]{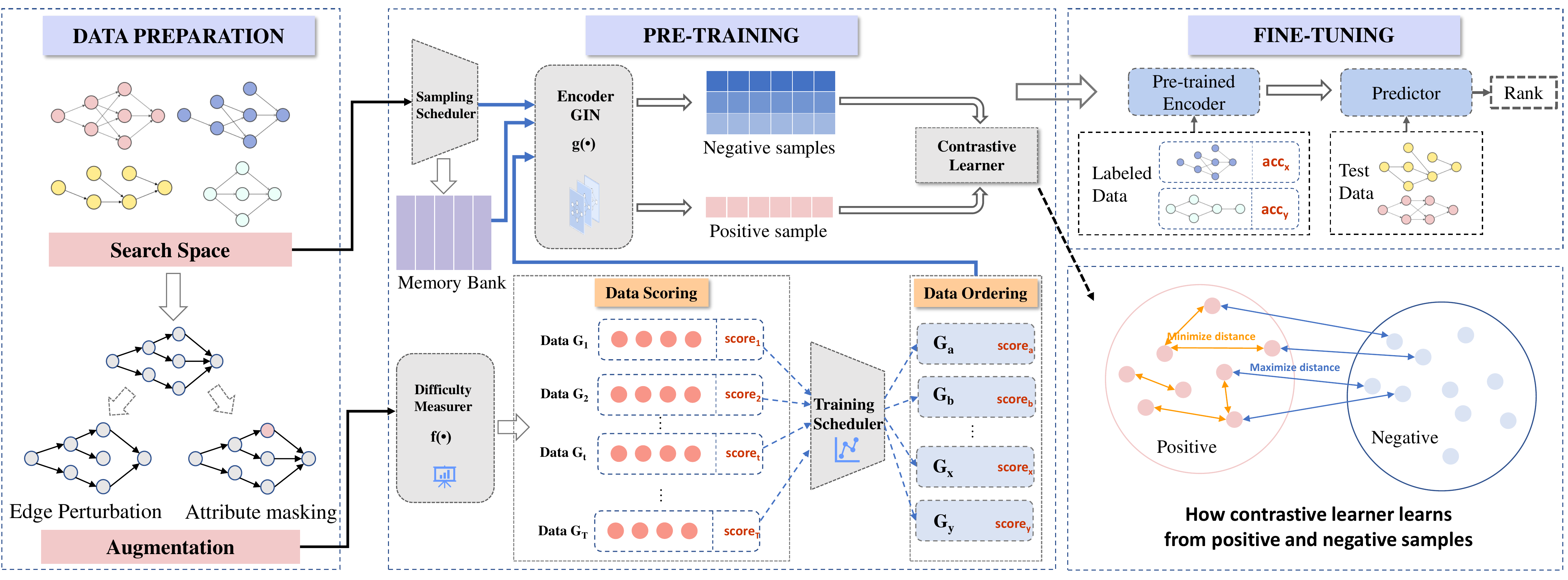}  
\caption{The overall framework of DCLP, comprises data preparation, pre-training, and fine-tuning. Left: The data preparation, uses edge perturbation and attribute masking to augment graphs in the search space to obtain the positive training samples. Middle: The pre-training. The bottom is the curriculum learning section, which schedules the training order of each positive data. The upper part is the contrastive learning process. A good encoder for neural networks is obtained by pre-training. Right: The upper part is the fine-tuning process, using a small amount of labeled data to fine-tune the encoder and obtain the predictor.}  
\label{figure:framwork}  
\end{figure*} 

To simplify the contrastive task for better adaptation to neural predictors, the issue of high variability in the distributions of randomly selected positive data for contrastive training, which may hinder the learning process for weak learners, must be addressed. This variability in data distribution arises from differences in the difficulty of positive samples during contrastive training. Specifically, when positive data are more similar to each other, the learner can classify them into the same class more easily, making such samples less challenging for contrastive learning. Then, our goal is to construct a training method with relatively stable data difficulty variation, allowing the contrastive learner to converge more consistently. To achieve this, inspired by curriculum learning~\cite{bengio2009curriculum}, we incorporate human knowledge to design a smooth learning curve, which enables the encoder to learn efficiently from unlabeled data. We propose a curriculum-guided contrastive learning method for neural predictors called DCLP as shown in Figure~\ref{figure:framwork}.

In DCLP, a novel metric is developed to measure the difficulty of positive data based on the distance between positive and original graphs. Next, a scheduler is employed to control the training order of data, prioritizing difficulty as the criterion as shown in the middle part of Figure~\ref{figure:framwork}. As training progresses, the scheduler gradually selects more difficult data, enabling the encoder to learn from harder data. However, unlike vanilla curriculum where data is learned from easy to hard, in DCLP, there are intervals of decreasing difficulty within the overall increasing trend. The decreasing intervals are arranged because as the data difficulty increases, the noise increases, and the model may overfit the noise, leading to poor performance. The uninterrupted use of simple data corrects the bias caused by noise to make the training more efficient. The scheduler infuses human knowledge into training, leading to efficient convergence of contrastive encoders, minimizing the empirical risk of training, and improving generalization ability ~\cite{wang2021survey}.

The key insight of DCLP is the difficulty of training varies across data, and mixing multiple difficulties in the same training round exacerbates the learning difficulty. Thus, the stepwise curriculum aims to improve the performance of predictors by creating a smooth training schedule.

After contrastive training, we use limited labeled data for fine-tuning. However, the typical regression loss is too stringent for NAS, which only requires the accurate predicted ranking~\cite{wang2021rank}. Then, we propose using ranking as the optimization target. Combining DCLP with various search strategies for NAS, our pre-training method employs designed learning curves to enhance contrastive tasks, enabling smooth extraction of valuable information from unlabeled data for prediction.  Additionally, we demonstrate the good generalization ability of DCLP to improve NAS performance while reducing consumption under multiple search strategies. Our contributions are summarized as follows:
\begin{enumerate}[labelsep = .5em, leftmargin = 0pt, itemindent = 1em]
    \item[$\bullet$] We utilize contrastive learning in neural predictors to leverage unlabeled data. This approach reduces the requirement for labeled training data of the predictor and improves its generalization ability.
    \item[$\bullet$] We propose a novel curriculum method to guide the contrastive task, which makes the predictor converge faster and perform better in NAS.
    \item[$\bullet$] We conduct comprehensive experiments, showing that DCLP outperforms the popular predictors on various benchmarks and search spaces that can be used in actual scenarios.
\end{enumerate}
\section{Related Work}
\noindent\textbf{Neural Architecture Search (NAS).} For NAS, the focus is primarily on search methods such as reinforcement learning~\cite{zoph2017neural}, evolutionary search~\cite{huang2022greedynasv2}, and random search~\cite{Yang2020NAS}. However, a challenge is the costly evaluation. Although the one-shot method utilizes super-networks to address the issue, it remains expensive~\cite{liu2018darts}. Our focus is on the neural predictor, an efficient method that utilizes predictive models to decrease estimation costs.

\noindent\textbf{Architecture performance predictors.} The predictor is a method for reducing the evaluation cost by directly estimating neural networks. PNAS~\cite{cai2018proxylessnas} introduces LSTM-based predictors. Other methods include random forests~\cite{liu2021homogeneous} and GNNs~\cite{jing2020self}. However, they only use labeled data, while using unlabeled data can reduce the cost of getting data. For instance, Semi-Supervised Assessor~\cite{tang2020semi} utilizes both labeled and unlabeled data, but the prediction is not ideal. Instead, DCLP makes unsupervised training more stable through curriculums. GMAE~\cite{ijcai2022p432} uses reconstruction as the pretext task. The difference is that DCLP uses contrastive training, achieving better results efficiently. Specifically, CTNAS ~\cite{chen2021contrastive} predicts the relationship between two neural networks. Although it uses contrastive learning and curriculum, it differs greatly from DCLP. While contrastive learning in DCLP is unsupervised, CTNAS refers to its supervised training as contrastive. Additionally, CTNAS uses the curriculum to update results, whereas our curriculum schedules stable training. Thus, there are fundamental differences between CTNAS and DCLP. More details are available in Appendix G.

\noindent\textbf{Contrastive learning.} Contrastive learning is an unsupervised method that has achieved success in various domains such as CV ~\cite{he2020momentum} and NLP~\cite{devlin-etal-2019-bert}. It aims to train encoders to obtain embeddings for data by minimizing the distance between positive items (similar data) while maximizing the distance between negative data.

\noindent\textbf{Curriculum learning.} Curriculum learning is a training method that orders data to simulate human learning~\cite{bengio2009curriculum}, and it has been shown to enhance generalization ability in various domains, such as CV~\cite{pentina2015curriculum} and NLP~\cite{platanios-etal-2019-competence}. In this study, we integrate it into neural predictors and combine it with contrastive learning to leverage its benefits.

\section{Methodology}
In this section, we present DCLP, a contrastive neural predictor with curriculum guidance. Figure~\ref{figure:framwork} provides an overview of DCLP, including the contrastive learning in Sec.~\ref{sec:contrastive learning}, and the curriculum used for training in Sec.~\ref{sec:curriculum learning}. Additionally, we discuss fine-tuning in Sec.~\ref{sec:fine-tuning}, and its potential use for NAS in Sec.~\ref{sec:search method}.

Before delving into the specifics, it is necessary to define the search space. We use cells as the search unit in line with most NAS works~\cite{zoph2017neural}. Each architecture is composed of L-stacked cells. Each cell can be represented as a DAG $G$. Cells have two formats: operation on nodes (OON) and operation on edges (OOE). OON is used as the default. In an OON cell, each node represents an operation, and the edge $e_{i\rightarrow j}$ indicates the information flow from node $i$ to $j$. Details are illustrated in Appendix B.
\subsection{Contrastive Learning Framework}\label{sec:contrastive learning}
The purpose of contrastive learning is to train an encoder for feature extraction from unlabeled neural networks. This subsection introduces data augmentation and the encoder, followed by the contrastive training approach.

\noindent\textbf{Data Augmentation.} Augmentation is employed to generate useful data to train a robust encoder. Given a graph $G$, we create an augmented graph $\hat{G}$ using a predefined distribution $p_{\theta,\mu}(G)$, where $\theta$ represents the augmentation method and $\mu$ is the change ratio. To ensure appropriate augmentation for graphs, we use edge perturbation and attribute masking~\cite{wu2021self}. Edge perturbation randomly edits edges, while attribute masking modifies node properties. In DCLP, positive data are generated by augmentation, while negative data are obtained by random sampling.

\noindent\textbf{Encoder Method.} Encoder is used to convert neural networks into embeddings. In contrastive learning, similar data should have similar embeddings. Therefore, we need a model that can distinguish between graphs, so we employ Graph Isomorphism Network (GIN)~\cite{xu2018how}, as it is highly capable of identifying differences in graphs. The propagation function of the $k$-th layer of a GIN is as follows:
\begin{equation}
h_n^k=MLP^k(h_n^{k-1}+\sum\nolimits_{n'\in N(n)}h_{n'}^{k-1})\label{equation:gin}
\end{equation}
where $h_n^k$ is the encoding of node $v_n$ in the $k$-th layer, and $N(n)$ is the set of node adjacent to $v_n$. Since a graph feature is required, the features can be integrated as follows:
\begin{equation}
h_G=concat(readout(\{h_n^k|n \in G\})|k=0,...,K)\label{equation:gin_readout}
\end{equation}

\noindent\textbf{Contrastive Training.} Contrastive learning trains the encoder by letting it determine data similarity and ensure close embeddings for positive data. Then, the similarity between networks $G_i$ and $G_j$ is measured using Radial Basis Function (RBF) with encoder representations denoted as $z(G)$: 
\begin{equation}
s(G_i,G_j)=exp(-\frac{d(z(G_i),z(G_j))}{2\sigma^2})
\end{equation}
where $d$ is distance measurement such as Euclidean distance. RBF maps items into a high-dimensional space to separate data that are hard to separate in the original space.

The contrastive task requires similar data to yield a larger $s$. InfoNCE is used as the optimized target. It is applied to positive pairs $\left\{q,k_+\right\}$ and negative data $\left \{ k_i\right\}_{i=1}^N$ as follows:
\begin{equation}
L=-log\frac{exp(s(q \cdot k_+)/ \tau)}{s(q \cdot k_+)/ \tau+\sum_{i=1}^Nexp(s(q \cdot k_i)/ \tau))} \label{equation:contrastive loss}
\end{equation}
where $\tau$ is a temperature parameter. Minimizing Eq.\ref{equation:contrastive loss} allows the encoder to extract features from unlabeled data by ensuring higher score of positive data than negative. Then, the encoder can reasonably encode networks for prediction.

\subsection{Curriculum learning}\label{sec:curriculum learning}
Training predictors using contrastive learning pose challenges when determining the similarity of neural networks compared to regular data. Random data ordering in existing contrastive methods may lead to convergence issues due to varying data distributions~\cite{chu2021cuco}. To address this, we propose a novel curriculum that schedules data to create a smooth learning curve, simplifying the contrastive task and enabling stable learning without training crashes. The curriculum involves two steps: difficulty measurement and training scheduler, which we will discuss in detail.
\begin{algorithm}[tb]
    \caption{Training procedure of DCLP}
    \label{alg:algorithm1}
    \begin{algorithmic}[1] %[1] enables line numbers
        \REQUIRE Training set $\mathcal{D}$,
        the number of iterations  $\mathcal{T}$,
        the augmentation distribution $p$,
        GIN encoder $f$
        \STATE \textbf{pre-training}
        \STATE Initialize parameters $\alpha$ of $f$ with kaiming initialization
        \FOR{$t=1$ to $\mathcal{T}$}
                \STATE $G \leftarrow RandomSample(\mathcal{D})$  \STATE Initialize $\hat G=\emptyset$
                \FOR{$i=1$ to $N$}
                    \STATE $\hat G_i \sim p_\theta(\hat G|G)$
                    \STATE $\hat G \leftarrow \hat G \cup \hat G_i $
                \ENDFOR
                \STATE $G_t \leftarrow argmax(\mathcal{P}_t(\hat G_i))_{i=1}^N$
                \STATE $z \leftarrow f(G)$, $z^{\prime} \leftarrow f(G_t)$
                \STATE $f_\alpha \leftarrow f_\alpha - \nabla_{f_\alpha}L_{NCE}$
        \ENDFOR
        \STATE \textbf{fine-tuning}
        \STATE $S_F \leftarrow LimitedRandomSample(D)$
        \STATE $L \leftarrow SelectedLossFunction(S_F)$
        % \STATE $\widetilde S_F \leftarrow Norm(S_F)$
        \FOR{$t=1$ to $\mathcal{T}$}
        \STATE $f_\alpha \leftarrow f_\alpha - \nabla_{f_\alpha}L$
        \ENDFOR
        \ENSURE  the trained predictor $f_\alpha$
    \end{algorithmic}
\end{algorithm}

\noindent\textbf{Difficulty Measurement.}
The difficulty function evaluates the difficulty of positive items in the contrastive task that aims to classify similar samples into the same class. Difficult positive data are those less similar to the original data, potentially causing misclassification by the contrastive learner. Therefore, the difficulty of augmented data is calculated based on their similarity to the origin. The higher the similarity, the lower the difficulty. To calculate the difficulty of a positive graph $\hat G$ related to the original graph $G$, we use the Wasserstein Distance(WD). WD measures the variability between the two graphs, which is a reasonable measure of difficulty. For practicality, the augmentation distribution can be considered uniform. Additionally, we approximate WD using Levenshtein Distance (LD) as follows:
\begin{equation}
\begin{split}
L(\hat G,G)&=(\int |p^{-1}_{\theta,\mu}(\hat G|G)-p^{-1}(G)dt)^{1/p}\\
% &=\int_{\mathbb{R}}|p_{\theta,\hat G}(t)-p_{\theta,G}(t)|dt\\
% &=\int_{\mathbb{R}}|U_{\theta,\mu,\hat G}(t)-U_{G}(t)|dt\\
& \approx \int_{\mathbb{R}}LD(\hat G_t,G_t,t)dt \\
&= cLD(\hat G,G) \label{equation:difficulty}
\end{split}
\end{equation}

The reasonableness of the approximation lies in that we need to calculate the distance between two uniform distributions denoted by $\mu$. Each point's distance can be viewed as an approximation of the LD between the graphs generated by related parameters at that point. By integrating these approximations, we obtain an approximation of the final LD. Since the difficulty value in Eq.5 can be approximated as a multiple of LD, and c is a constant, the WD between the two graphs can be directly approximated by LD. This expresses the similarity between networks in terms of their structural similarity. More details can be found in Appendix F.1. Then the difficulty of $\hat G$ can be defined as:
\begin{equation}
L(G,\hat G)=\frac{LD(G,\hat G)}{|G|\cdot |\hat G|}
\end{equation}
where $|G|$ is the size of $G$ measuring by the number of edges.

\noindent\textbf{Training Scheduler.}
After defining difficulty, we design the training order based on difficulty, controlled by a parameter that determines the training preference for easy or hard data. The probability of sampling $\hat G_i$ at step t is then defined as:
\begin{equation}
\mathcal{P}_t(\hat G_i)=\frac{exp(\tau_t L(\hat G_i,G_i))}{\sum_{j=1}^nexp(\tau_t L(\hat G_j,G_j))}
\end{equation}
where
\begin{equation}
\begin{aligned}
&\tau_t=\frac{\tau_T-\tau_m}{\sigma}g(\sigma_t)+\tau_m,\tau_m =\frac{\tau_1+\tau_T}{2}\\
&\sigma_t=g^{-1}(\frac{t}{T}\sigma+\frac{\sigma}{k}sin\frac{n\pi t}{T})
\end{aligned}
\end{equation}
where $t$ denotes steps ($t \in \{1,2,...T\}$), $g$ is tanh, $\sigma=0.9$, and $n>1, k>1$ controls the frequency and range of difficulty decrease respectively, the decreasing will be discussed in the next paragraph. $\tau_t$ determines the preference for easy/hard data. For $\tau_t<0$, larger difficulty corresponds to a smaller selection probability. Conversely, when $\tau_t>0$, the curriculum favors hard data. Our strategy gradually adjusts $\tau_t$ to achieve robust learning. Initially, we set a negative $\tau_1$ to emphasize learning from easy data at the start without being affected by noise and to quickly reach a relatively good ability. Then we gradually increase $\tau_t$ to encourage deeper learning from hard data. Finally, we use a positive $\tau_T$, smoothly transitioning from easy to hard data.

However, unlike the conventional easy-to-hard curriculum, we introduce a fluctuating term that causes the curriculum to fluctuate in difficulty controlled by $n$ and $k$. This non-monotonic increase in difficulty includes intervals of decreasing difficulty. Therefore, the model can revisit simpler data periodically during training. It is necessary since the harder positive data includes more noise, i.e., negative samples that are mistakenly labeled as positive. Training with increasing difficulty often causes the model to focus excessively on the noise, resulting in loss of learned knowledge in the correct data. Intermittent training with simple data can counteract the learned misinformation from the noise, improving training stability, and being more robust.

The curriculum facilitates the encoder to acquire a stronger ability by increasing data difficulty. Additionally, incorporating easy data prevents overfitting that may occur as difficulty increases, leading to efficient convergence.

As shown in Algorithm~\ref{alg:algorithm1}, we use the difficulty function to calculate the difficulty of each positive data and sort them according to the scheduler (Line 10). Next, we feed the positive samples into the contrastive learner for training.

\subsection{Fine tuning}\label{sec:fine-tuning}
After pre-training, to transfer the pre-trained model to the prediction task, fine-tuning is in demand. However, using mean squared error (MSE) between predicted and actual results as loss function is too strict, as NAS focuses on the best networks. To alleviate the issue of overfitting, we propose two functions that take ranking into account.

\noindent\textbf{MSE+normalization.} As we prioritize ranking, normalization can widen the performance gap between two networks. Inaccurate rank judgments lead to greater MSE, facilitating the learning of their relative ranking relationship. Further details can be found in the supporting material F.2.

\noindent\textbf{Ranking loss.} We utilize ListMLE~\cite{burges2010ranknet} to optimize predicted ranking. We use $\it o$ to represent the actual sorting result and $S=\left\{s_1...s_n\right\}$ to denote the predicted accuracy of $n$ architectures. Then, $S$ is sorted according to $\it o$ to obtain $\hat S=\left\{s_{o_1},s_{o_2},... .s_{o_n}\right\}$. Our objective is to maximize the probability of this $\hat S$ to make the predicted ranking similar to the actual one. The loss function is as follows:
\begin{equation}
L=-\sum_{i=1}^n log\frac{exp(s_{o_i})}{\sum_{j=i}^nexp(s_{o_j})}
\end{equation}

These two optimization targets align the optimization goal with the prediction task, improving the model’s fitness to NAS while reducing the required labeled data. 
\begin{algorithm}[tb]
    \caption{Predictor-based Random Search Strategy}
    \label{alg:algorithm2}
    \begin{algorithmic}[1] %[1] enables line numbers
        \REQUIRE Search space $\mathcal{A}$, 
        predictor $f$,
        iteration steps $\mathcal{T}$,
        the number of networks sampled in \textit{t}-th iteration $\mathcal{N}^t$,
        the number of networks selected in each iteration $\mathcal{K}$.
        \STATE Initialize $t=0$, an architecture set $S=\emptyset$
        \FOR{$t=1$ to $\mathcal{T}$}
        \STATE Randomly sample $\mathcal{N}^t$ architectures $S^t=\left\{a_j^t\right\}_{j=1}^{\mathcal{N}^t}$ from the search space $\mathcal{A}$ without repeat.
        \STATE Evaluate items in $S^t$ by trained predictor $f$, get $\hat S^t=\left\{(a_j^t,y_j^t)\right\}_{j=1}^{\mathcal{N}^t}$,where y is the 
        predicted performance.
        \STATE Select top-k items from $\hat S^t$ according to y, get $\widetilde S^t$. 
        \STATE $S \leftarrow S \cup \widetilde S^t$.
        \ENDFOR
        \STATE Evaluate each architecture in S by training to obtain the corresponding ground-truth $\mathcal{Y}$
        \ENSURE  the architecture $a^*$ with the best true performance.
    \end{algorithmic}
\end{algorithm}
\subsection{Search method}\label{sec:search method}
After fine-tuning, we use DCLP with various search strategies, such as random search (RS), evolution search (EA), and reinforcement learning (RL), to discover the optimal network. The fast estimation provided by DCLP aids in the efficient exploration of search space. A larger exploration space increases the likelihood of finding a better result.

Random search is outlined in Algorithm~\ref{alg:algorithm2}, while other strategies are in Appendix E. Each iteration involves randomly sampling $N$ networks, and the top-$k$ based on predicted performance are added to the candidate pool (Lines 4-6). After $T$ iterations, only a few networks ($k \cdot T$) are trained to obtain actual performance, from which the optimal network is selected, making it an efficient approach.

\section{Experiments}
\begin{figure}  
\centering  
\includegraphics[height=5.4cm,width=8cm]{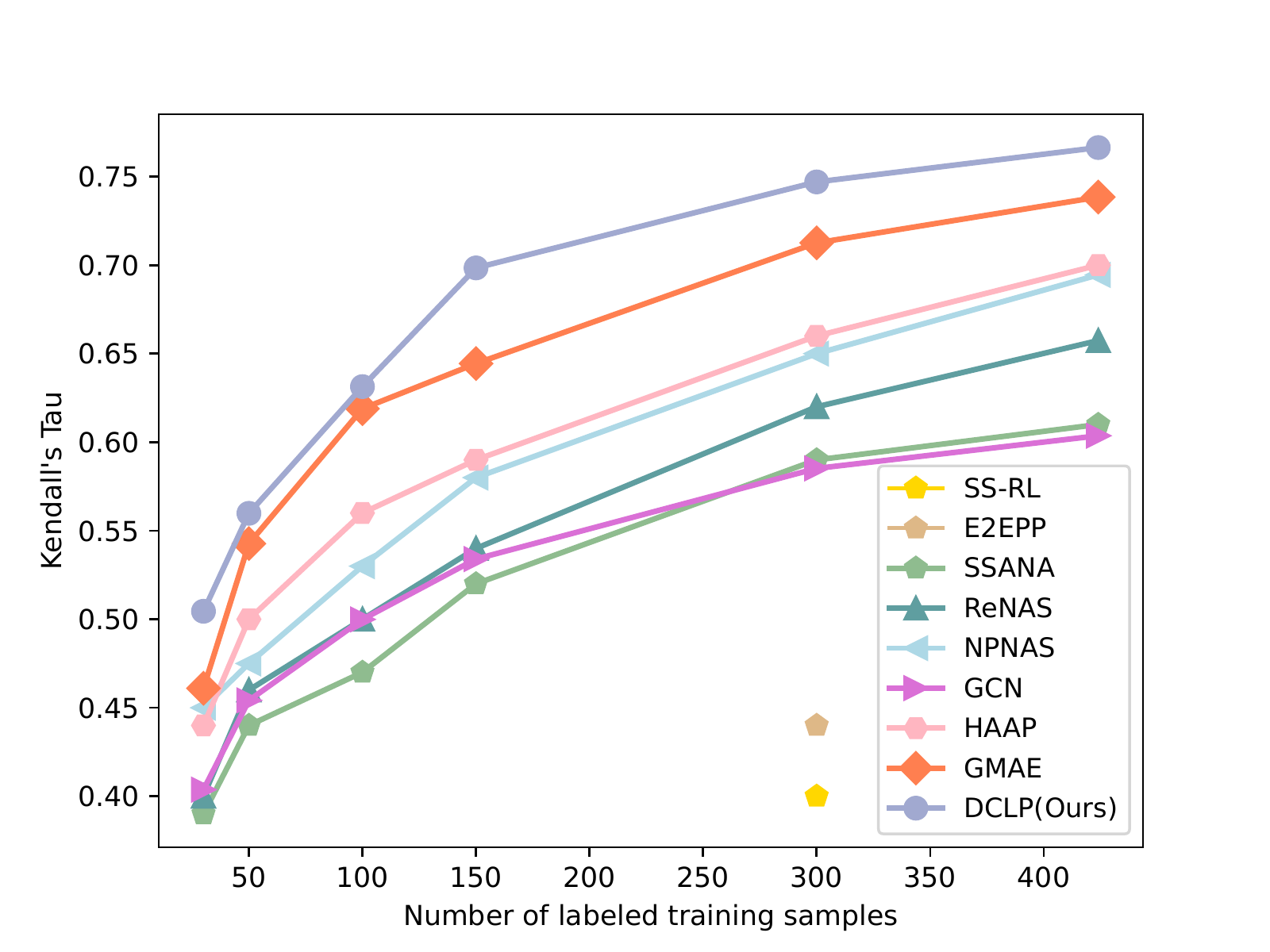}  
\caption{The comparison between DCLP and the state-of-the-art predictors on NAS-Bench-101.} 
\label{figure:nas-bench-101 ktau}  
\end{figure} 
We evaluate DCLP through experiments for image classification and language tasks. We introduce experimental settings first and validate the superiority in NAS benchmarks,  along with ablation experiments for further analysis.

\noindent\textbf{Image classification.} We conduct experiments on two NAS benchmarks and a large search space. Details are as follows.
\begin{enumerate}[labelsep = .5em, leftmargin = 0pt, itemindent = 1em]
    \item[$\bullet$]\textbf{NAS-Bench-101.} It is an OON set of 423k CNN~\cite{ying2019bench}. Each node can be a 1x1 convolution, etc. Test accuracy on CIFAR-10 is available within the search space.
    \item[$\bullet$]\textbf{NAS-Bench-201.} This is a NAS benchmark in OOE space~\cite{Dong2020NAS-Bench-201:} with 15,625 structures, providing performance on CIFAR-10, CIFAR-100, and ImageNet. 
     \item[$\bullet$]\textbf{DARTS Space.} It is an OOE search space ~\cite{liu2018darts}, consisting of $10^{18}$ architectures. It contains cells with each edge offering an operation choice.
\end{enumerate}
\noindent\textbf{Language model.}
We demonstrate the efficiency of DCLP on the language model task. It typically uses RNN instead of CNN in image tasks. The success of DCLP on both types of neural networks shows its generalization capability.
\begin{enumerate}[labelsep = .5em, leftmargin = 0pt, itemindent = 1em]
    \item[$\bullet$]\textbf{DARTS RNN Space:} The recurrent cell is an OOE space. The operations available are ReLU, Tanh, etc.
\end{enumerate}

\noindent\textbf{Evaluation Metrics.} We evaluate the predictor using Kendall's Tau~\cite{sen1968estimates} for ranking accuracy, with a score closer to 1 indicating a better predictor. Additionally, the image classification accuracy and perplexity for language models are used to measure the performance of the NAS method. 

\begin{table*}[htbp]
  \centering
    \small
    \begin{tabular}{c|r|r|r|r|r|r|c|c|c}
    % \toprule
    \Xhline{1pt}
    \multirow{2}{*}{Method} & \multicolumn{2}{c|}{CIFAR-10} & \multicolumn{2}{c|}{CIFAR-100} & \multicolumn{2}{c|}{ImageNet-16-120}& \multirow{2}{*}{\makecell[c]{Query}}& \multirow{2}{*}{\makecell[c]{Search cost\\(GPU h)}} & \multirow{2}{*}{Type} \\
    \cline{2-7}
          & \multicolumn{1}{c|}{Acc(\%)} & \multicolumn{1}{c|}{R(\%)} & \multicolumn{1}{c|}{Acc(\%)} & \multicolumn{1}{c|}{R(\%)} & \multicolumn{1}{c|}{Acc(\%)} & \multicolumn{1}{c|}{R(\%)} & & \\
    \Xhline{0.4pt}
    % DARTS~\cite{liu2018darts}  & 54.30±0.00 & 99.6 & 15.61±0.00 & 96.18 & 16.32±0.00 & 91.19 & - & 8.3& supernet\\
    % GDAS~\cite{dong2019searching}  & 93.51±0.13 & 2.63 & 70.61±0.26 &2.05 & 41.84±0.90 & 15.21 & - &8.0& supernet\\
    NASWOT~\shortcite{mellor2021neural}  & 93.15±0.08 & 7.0 & 69.98±1.22 & 5.01 & 44.44±2.10 & 3.01 & - &0.06& zero+RS\\
    TENAS~\shortcite{chen2020tenas}  & 93.18±0.39 & 6.6 & 70.24±1.05 & 3.68 & 43.55±8.24 &6.6 & - &0.43& zero\\
    HAAP~\shortcite{liu2021homogeneous}  & 93.75±0.17 & 1.16 & 71.08±0.19 & 0.97 & 45.22±0.26 & 0.94 & 150 &0.011&pred+EA\\
    % TF-MOENAS~\cite{do2021training}  & 93.96±0.10 & 0.31 & 71.94±0.50 & 0.34 & 45.68±1.24 & 0.40 & - &2.5& zero+EA\\
    GMAE~\shortcite{ijcai2022p432} & 94.03±0.11 & 0.21 & 72.56 ±0.16 & 0.13 & 46.09±0.27 & 0.19 & 150 &0.009&pred+RS\\
    HOP~\shortcite{chen2021not}  & 94.1±0.12 & 0.14 & 72.64± 0.11 & 0.10 & 46.29±0.19 & 0.1 & 150 &0.015 &pred+RS\\
    
    FreeREA~\shortcite{cavagnero2023freerea}  & 94.20±0.02 & 0.08 & 73.30±0.31 & 0.01 & 46.34±0.00 & 0.09 & 100 &0.012& EA\\
    DCNAS~\shortcite{pan2022distribution}  & 94.29±0.07 & 0.14 & 73.02±0.16 & 0.02 & 46.41±0.14 & 0.08 & - &3.9& supernet\\
    \Xhline{0.4pt}
    DCLP+RL  & 94.29±0.05 & 0.14 & 72.83±0.15 & 0.096 & 46.49±0.25 & 0.03 & 50&0.007 &pred+RL\\
    DCLP+RS  & \textbf{94.34}±0.03 & \textbf{0.025} & \textbf{73.5}±0.3 & \textbf{0.0001} & \textbf{46.54}±0.03 & \textbf{0.02} & 50&0.004&pred+RS\\
    \Xhline{1pt}
    \end{tabular}%
    
  \caption{Comparison with the SOTA methods on NAS-Bench-201. Query indicates the amount of labeled data needed for training the predictor, the less the better. R means ranking, pred means predictor, and zero means zero-cost estimation proxy.}
  \label{tab:4}%
\end{table*}%
\begin{table}[htbp]
  \centering
  \small
    \begin{tabular}{crrrrr}
    \toprule
    Method & Accuracy(\%) &  Query & Type\\
    \midrule
    % DARTS~\cite{liu2018darts} & 92.21±0.61& 19.64 & -- & gradient\\
    % FBNetV3\cite{dai2020fbnetv3} & 92.29±1.25&  --& gradient\\
    HAAP~\shortcite{liu2021homogeneous}   & 93.69±0.22 & 300 & pred+EA\\
    FreeREA~\shortcite{cavagnero2023freerea}  & 93.80±0.02& 500& EA \\
    ReNAS~\shortcite{xu2021renas} & 93.90±0.21 &  423 & pred+EA\\
    CTNAS~\shortcite{chen2021contrastive} & 93.92±0.18 &  423 & pred+RS\\
    GMAE~\shortcite{ijcai2022p432}  & 93.98±0.15 & 300 & pred+EA\\
    HOP~\shortcite{chen2021not}   & 94.09±0.11 & 300 & pred+RS\\
    \midrule
    % DCLP+EA  & 94.09±0.06 &  300 &pred+EA\\
    DCLP+RL  & 94.14±0.08 &  300 &pred+RL\\
    DCLP+RS  & \textbf{94.17±0.06} & 300 &pred+RS\\
    \bottomrule
    \end{tabular}%
  \caption{The comparison of NAS on NAS-Bench-101. Query indicates the amount of labeled training data, the less the better, and pred indicates predictor.}
  \label{tab:3}%
\end{table}%

\subsection{Predictor Evaluation}\label{sec:exp/predictor evaluation}
In this section, we evaluate the predictive ability using Kendall's Tau on NAS-Bench-101. Predictors with high Kendall's Tau are more likely to discover exceptional networks. We compare DCLP with popular predictors~\cite{wei2021self,xu2021renas,ijcai2022p432,wen2020neural,liu2021homogeneous,kipf2017semisupervised} to demonstrate its superiority in requiring less labeled data while maintaining high performance. Our goal is to develop efficient predictors for NAS with minimal labeled data, and DCLP shows outstanding performance with small labeled sets.

As shown in Figure~\ref{figure:nas-bench-101 ktau}, DCLP outperforms various popular predictors, particularly with smaller training sets. This is consistent with our focus on achieving high performance on small training data as mentioned above.

The results demonstrate DCLP's superior generalization with small training sets compared to solely supervised learning approaches~\cite{xu2021renas,wen2020neural,kipf2017semisupervised}. This is consistent with our intention of reducing dependence on labeled data by using curriculum-guided contrastive learning to train the encoder. While methods relying on supervised learning focus on improving the predictor structure to extract more knowledge from limited labeled data, the amount of information remains limited, leading to poor performance. Hence, DCLP's advantage over these methods reinforces the effectiveness of contrastive pre-training in constructing robust neural predictors.

On the other hand, DCLP outperforms unsupervised predictors~\cite{wei2021self,ijcai2022p432,liu2021homogeneous}, validating the effectiveness of our curriculum-guided learning. The moderate learning and ranking of training data enhance the contrastive learning process, aligning with our motivation. Since our key point is the pre-training for neural predictors, extending this approach to other methods is natural. This extension is critical in constructing effective predictors that rely on limited trained architectures.

\subsection{Search Strategy Performance Evaluation}\label{sec:exp/search strategy}
\begin{table}[htbp]
  \centering
  \small
    \begin{tabular}{ccccc}
    \Xhline{1pt}
    \multirow{2}{*}{Method} & \multirow{2}{*}{\makecell[c]{Test error\\(\%)}} & \multicolumn{2}{c}{Cost(GPU h)}& \multirow{2}{*}{Type}\\
    \cline{3-4}
    & &train & search & \\
    \Xhline{0.4pt}
    DARTS-~\shortcite{chu2021darts} & 2.59±0.08 & -  & 9.6 & supernet\\
    CTNAS~\shortcite{chen2021contrastive}  & 2.59±0.04 &120   & 2.4& pred+RL\\
    % NOSH~\cite{wang2021rank}  & 2.53±0.02 & - & 48.2& RS\\
    % BANA~\cite{white2021bananas}  & 2.67±0.07 & - & 244.8& pred+BO\\
    CATE~\shortcite{yan2021cate} & 2.56±0.08 & - & 75& pred+RS\\
    HOP~\shortcite{chen2021not}  & 2.52±0.04 & - & 60& pred+RS\\
    GMAE~\shortcite{ijcai2022p432} & 2.50±0.03 & 40  & 0.2& pred+BO\\
    DCNAS~\shortcite{pan2022distribution}& 2.50±0.01 & -  & 480 & supernet\\
    \Xhline{0.4pt}
    DCLP-RL  & \textbf{2.50}±0.02 & 4.0
&  0.08& pred+RL\\
    DCLP-RS  & \textbf{2.48}±0.02 & 4.0
&  0.05& pred+RS\\
    \Xhline{1pt}
    \end{tabular}%

  \caption{The comparison of NAS in DARTS space with CIFAR-10 as the dataset. The pred means predictor.}
  \label{tab:DARTS-CNN}%
\end{table}
In this section, we compare DCLP with popular methods to demonstrate its effectiveness with search methods. We report our setups in Appendix D. As mentioned above, we aim to achieve the best results with the smallest labeled sets in the shortest time. Therefore, we introduce two evaluation metrics: \textit{Query} and \textit{Cost}. \textit{Query} represents the number of trained networks required for predictor, and smaller values indicate more efficient training. \textit{Cost} encompasses computational power consumed in training and search. For predictor-based methods, it includes the cost of training neural networks to obtain training data, along with the time for training the predictor and search process. The unit of cost is GPU hour (i.e. GPU h), which represents one hour of calculation on a GPU. We use a single RTX3090 as the platform.
\begin{figure*}
        % \vspace{-0.4cm}
	\centering
	\subfigure[With/without contrastive learning]{
		\begin{minipage}[t]{0.3\linewidth}
			\centering
			\includegraphics[width=1\textwidth]{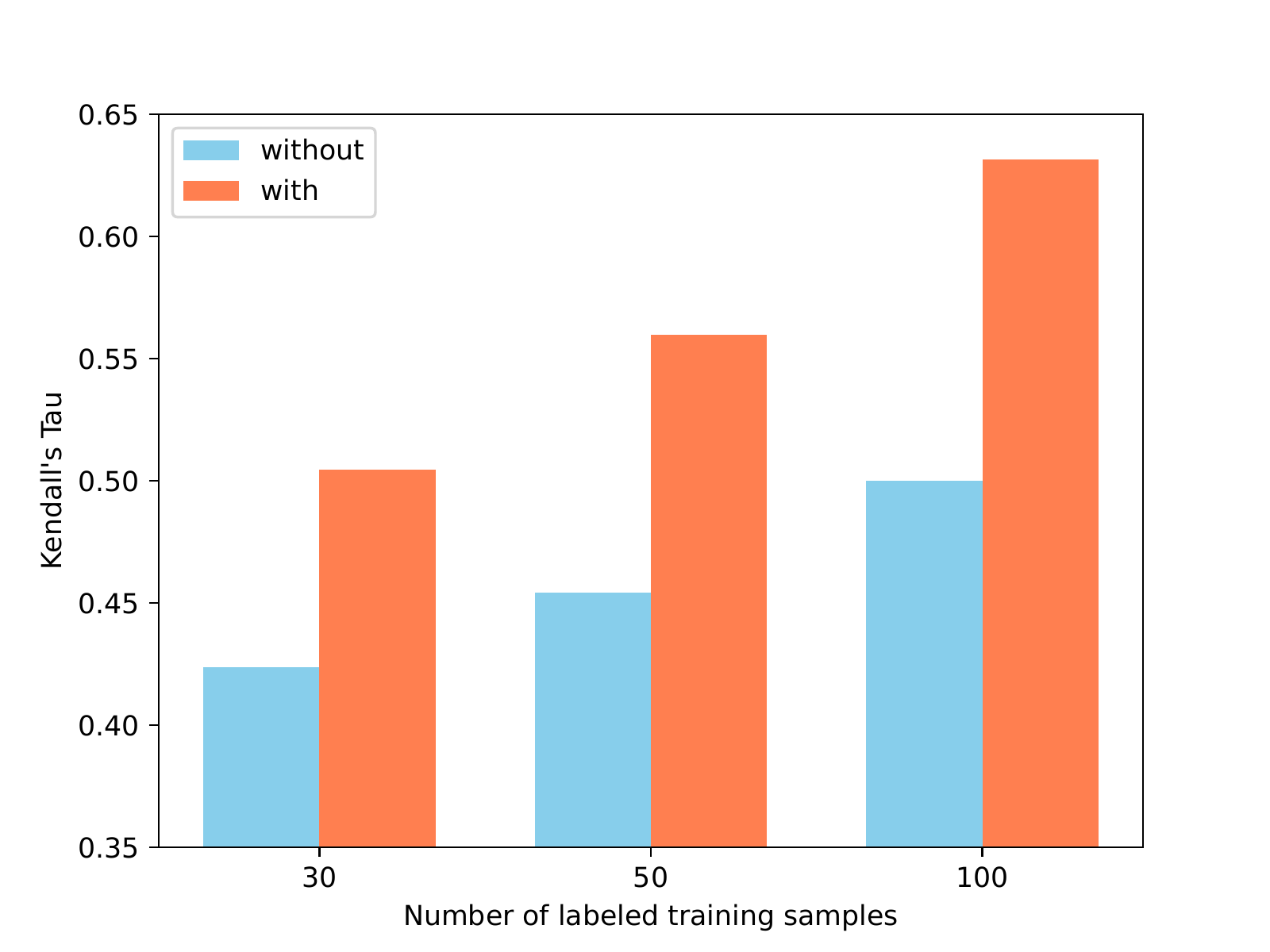}
		\end{minipage}
	}%
	\subfigure[Choice of curriculum learning]{
		\begin{minipage}[t]{0.3\linewidth}
			\centering
			\includegraphics[width=1\textwidth]{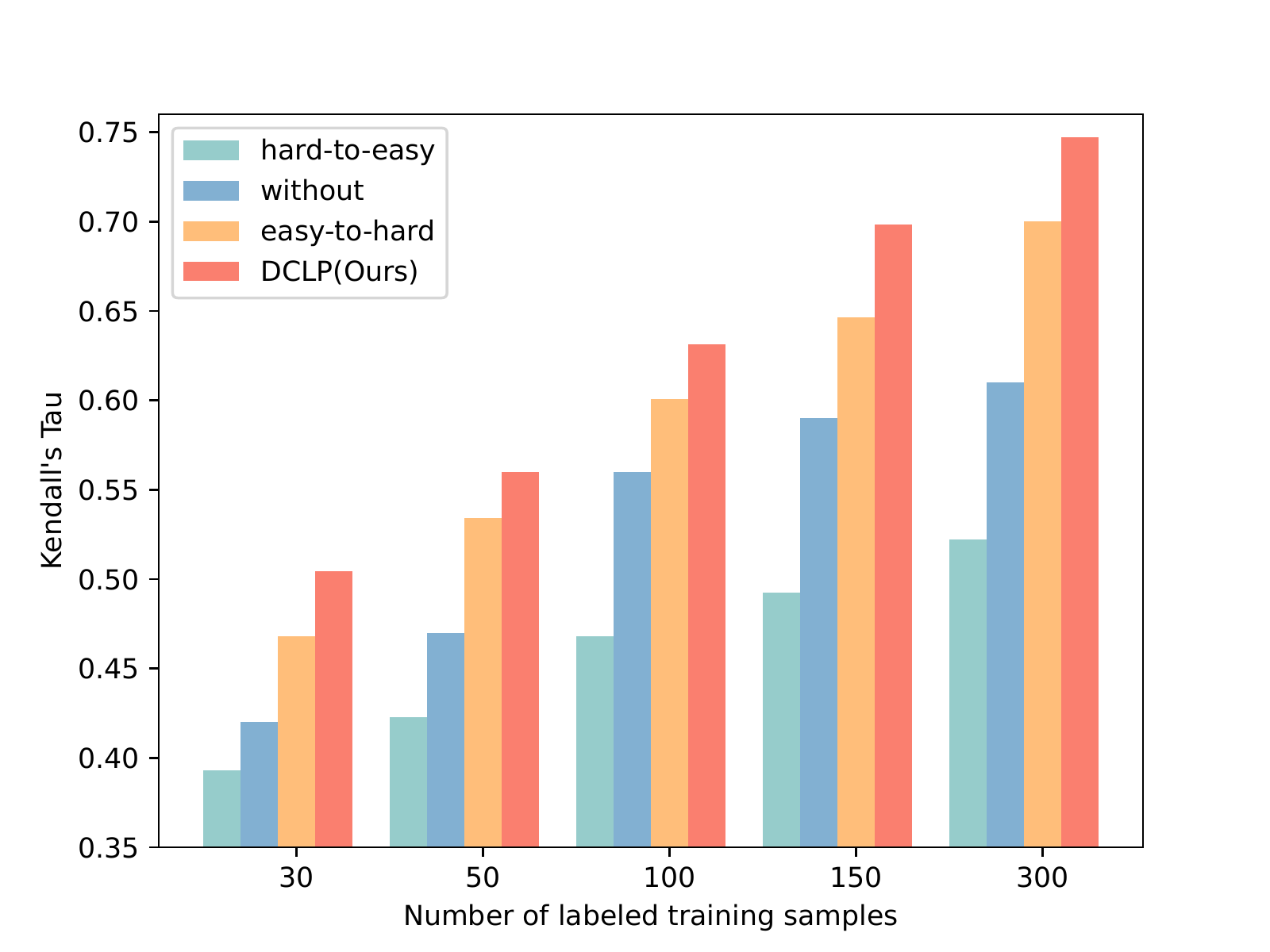}
		\end{minipage}
	}%
	\subfigure[Choice of fine-tuning methods]{
		\begin{minipage}[t]{0.3\linewidth}
			\centering
			\includegraphics[width=1\textwidth]{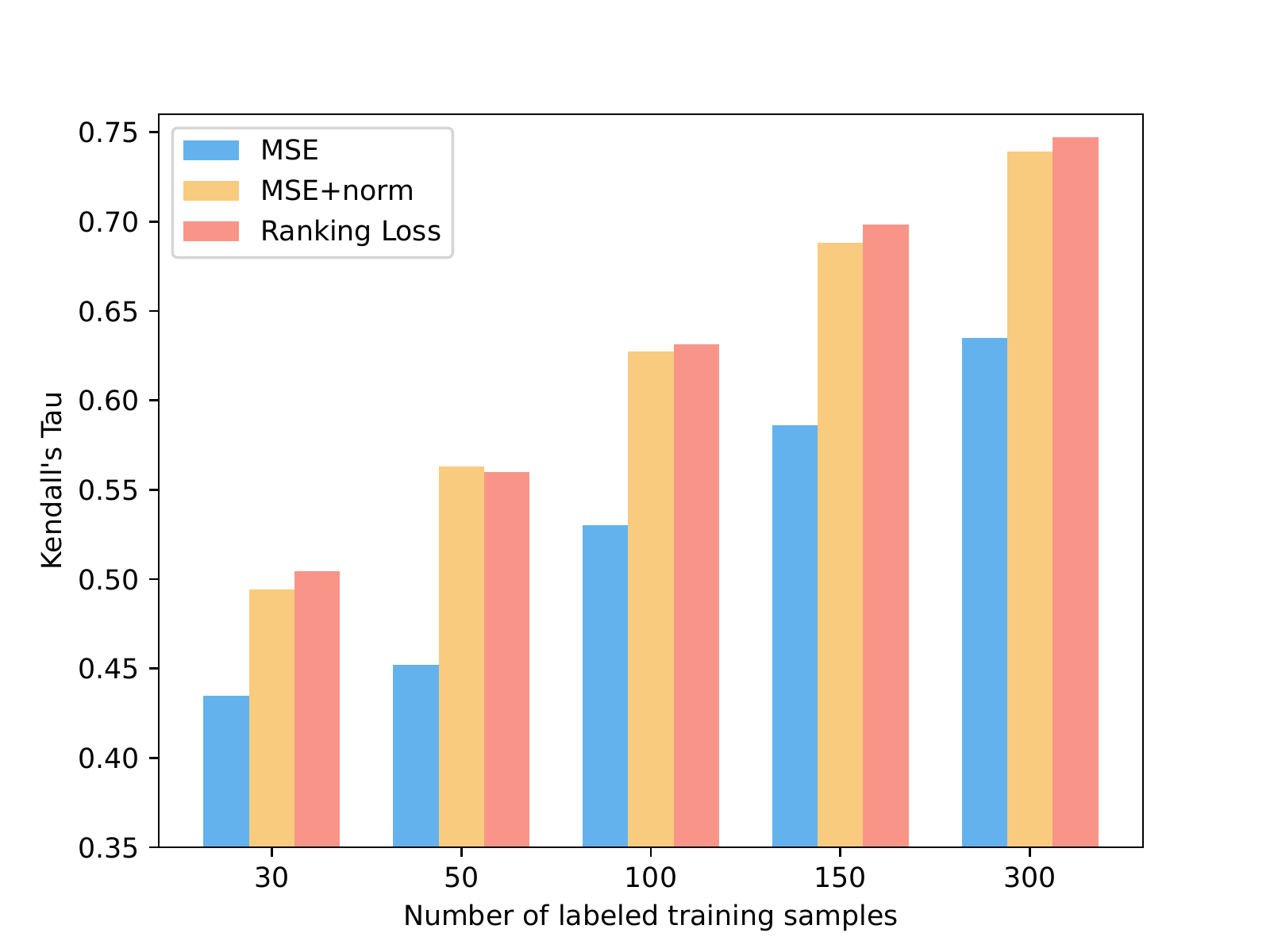}
		\end{minipage}
	}%
	\centering
     \vspace{-0.2cm}
	\caption{The ablation of contrastive learning, curriculum learning, and fine-tuning targets on NAS-Bench-101.}
	\label{ablation}
\end{figure*}

Tables~\ref{tab:4} and \ref{tab:3} present experiments on NAS-Bench-201 and NAS-Bench-101, respectively. DCLP achieves the highest precision, with minimal training data and cost. This is attributed to leveraging knowledge from unlabeled data effectively. Additionally, DCLP achieves better results through curriculum compared to the predictor which employs self-supervised or zero-cost methods. Moreover, it offers nearly zero-cost estimation, accelerating evaluation while maintaining superior accuracy compared to agent-dependent methods. Interestingly, combining DCLP with RS yields the best results as RS is a strong method~\cite{Yang2020NAS}, especially when the search range is large~\cite{bergstra2012random}. DCLP enables RS to explore a wide range effectively, enhancing its performance.
\begin{table}[htbp]
  \centering
  \small
    \begin{tabular}{ccccc}
    \Xhline{1pt}
    \multirow{2}{*}{Method} & \multirow{2}{*}{Test PPL} & \multicolumn{2}{c}{Cost(GPU h)}& \multirow{2}{*}{ Type}\\
    \cline{3-4}
    & &train & search& \\
    \Xhline{0.4pt}
    % DARTS~\cite{liu2018darts}  &58.37 & - & 5 & supernet\\
    DARTS-~\shortcite{chu2021darts}  & 59.21 & -  & 4.2& supernet\\
    DCNAS~\shortcite{pan2022distribution} & 58.21  & - & 3.2 & supernet\\
    $\beta$-DARTS~\shortcite{ye2022b} & 58.17 & - & 3.1 & supernet\\
   
    GAME~\shortcite{ijcai2022p432}  & 57.41 & 3.5 & 0.15 & pred+BO\\
    HOP~\shortcite{chen2021not}  & 57.34 & -  & 4.0& pred+RS\\
     CL-V2~\shortcite{zhou2022curriculum}  &  56.94 & - & 4.8& supernet\\
    \Xhline{0.4pt}
    DCLP-RS  & \textbf{56.19} & 2.0 & 0.04&pred+RS \\
    \Xhline{1pt}
    \end{tabular}%
  \caption{The comparison of NAS in DARTS space, and we use PTB as the dataset for the language model task. Pred means predictor, and zero means zero-cost estimation proxy.}
  \label{tab:RNN}%
\end{table}

Table~\ref{tab:DARTS-CNN} and~\ref{tab:RNN} present experimental results on DARTS space for image classification and language model tasks. More results on ImageNet are available in the Appendix. DCLP outperforms predictor-based and other evaluation strategies, showcasing its superiority in large search spaces with reduced training and search costs. Notably, DCLP achieves better results in less time compared to other methods, further validating its effectiveness. Additionally, DCLP's success in the large CNN and RNN search spaces suggests its potential for broader application to various neural structures, reinforcing the effectiveness of our curriculum-based contrastive training for neural predictors.

\subsection{Ablation Study}\label{sec:exp/ablation}
\noindent\textbf{With/Without contrastive learning.} In Figure~\ref{ablation}(a), we demonstrate that using contrastive learning for pre-training is superior to training the predictor from scratch. This supports the objective of using contrastive learning to acquire information about the features of neural networks through the contrastive task on unlabeled data. By leveraging this information, the labeled data needed for the prediction task's training process is substantially decreased, leading to resource savings while obtaining a better predictor.

\noindent\textbf{Choice of curriculum learning methods.} As shown in Figure~\ref{ablation}(b), we compare results without curriculum and using different curriculum methods, revealing a significant impact. Surprisingly, learning from hard to easy is worse than no curriculum. This is because determining the similarity of neural networks is challenging, even when simplified as graph isomorphism (NP-hard). This leads to an unclear boundary between difficult positive and negative samples so the difficult data will contain noise, i.e., negative samples are incorrectly labeled as positive. Then, if the training is too difficult at the beginning, the encoder tends to focus more on the noise that often appears in hard data, which degrades performance. On the other hand, learning without a curriculum is ineffective as the learner 
\begin{table}[htbp]
  \centering
    \begin{tabular}{cccccc}
    \Xhline{1pt}
    \multirow{2}{*}{\makecell[c]{Contrastive \\Learning}} & \multirow{2}{*}{\makecell[c]{Curriculum \\Learning}} & \multirow{2}{*}{\makecell[c]{Ranking \\Loss}} & \multirow{2}{*}{\makecell[c]{Test error\\(\%)}}\\
    & & & &  \\
    \Xhline{0.4pt}
    \ding{56}  & \ding{56} & \ding{56}  & \textbf{6.42}±2.00\\
    \ding{56}  & \ding{56} & \ding{52}  & \textbf{5.80}±1.75\\
    \ding{52}  & \ding{56} & \ding{52}  & \textbf{5.36}±0.95\\
    \ding{52}  & easy to hard & \ding{52} &\textbf{2.66}±0.17\\
    \Xhline{0.4pt}
    \ding{52}  & ours & \ding{52}& \textbf{2.48}±0.02\\
    \Xhline{1pt}
    \end{tabular}%
  \caption{The ablation experiment in DARTS space on CIFAR-10. If without pre-training, we use the time for pre-training to obtain more labeled data for supervised training.}
  \label{tab:ablation}%
\end{table}
struggles to converge with randomly selected data of varying difficulties. Tasks progressing from simple to hard yield excellent results, confirming the correctness of the overall trend. 
However, using complex data introduces noise, requiring some easy data for parameter optimization. Our method cycles between hard and easy tasks, avoiding noise exploration and ensuring the ability to learn complex data for improved performance.

\noindent\textbf{Choice of fine-tuning targets.} Figure~\ref{ablation}(c) shows the impact of different fine-tuning functions. The choice of target function significantly affects the predictor's performance. Notably, ranking loss and MSE+norm outperform vanilla MSE. This result aligns with our concept. The effectiveness of using ranking as the optimization target in this task is demonstrated. As ranking loss emphasizes the rankings, it shows particular advantages as training sets grow in size.

Table~\ref{tab:ablation} compares the effects of various techniques in DCLP using DARTS space. When pre-training is not used, we allocate the time to obtain more labeled training data. The results further support the significance of curriculum-guided contrastive learning for high-performance predictors.
\section{Conclusion}
In this paper, we propose DCLP, a curriculum-guided contrastive pre-training approach that improves neural predictor performance in NAS. By pre-training the encoder with difficulty-scheduled positive data, the training process becomes stable and efficient. Fine-tuning with a suitable target further improves the predictor. DCLP reduces reliance on labeled data, enabling efficient and accurate predictor-based NAS. It offers a promising approach for neural predictors in NAS and future work will focus on designing self-supervised pretext tasks to boost performance further.

\section{Acknowledgments}
\noindent This paper was supported by NSFC grant~(62232005, 62202126) and The National Key Research and Development Program of China~(2020YFB1006104).

\bibliography{aaai24}

\newpage

\section{Appendix}
\appendix

\section{Code \& Data}
The DCLP code and datasets are included as supplementary materials. Because of the page limit and the fact that the datasets used in this study are readily accessible from GitHub, we only provide links to open-source projects related to the datasets. As stated previously, when comparing our work with others, we utilize the code and parameters from their open-source projects.
\section{Architecture of Neural Cell}
\begin{figure}  
\centering  
\includegraphics[height=2.2cm,width=8.5cm]{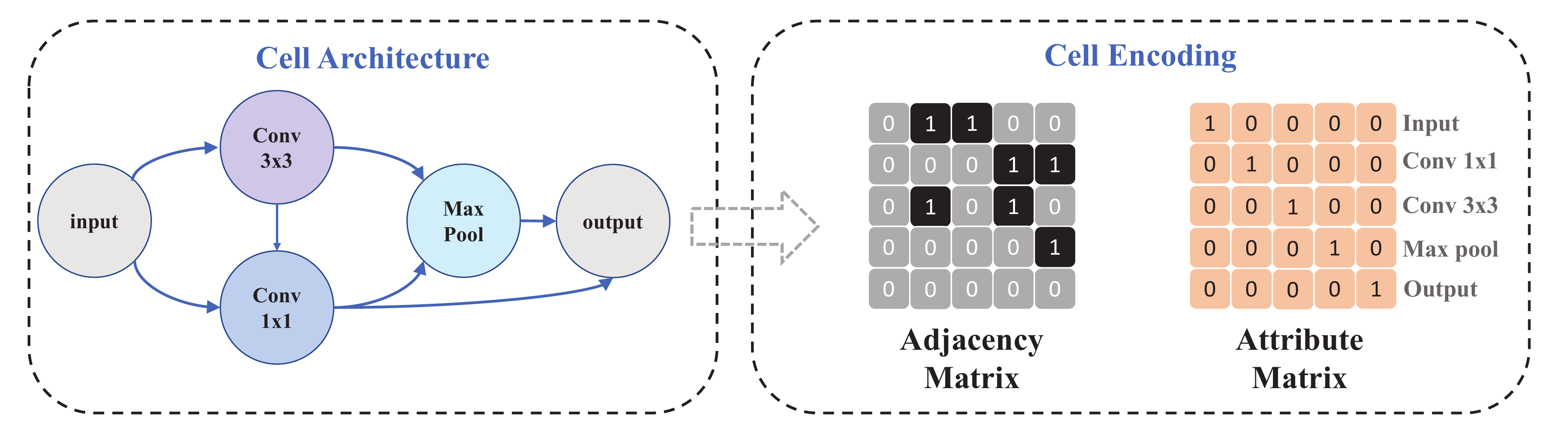}  
\caption{The DAG and matrixes correspond to a neural cell.}  
\label{figure:cell}  
\end{figure}

In Section 3, we have explained that each neural architecture cell can be depicted as a directed acyclic graph (DAG), and this DAG can be represented through an adjacency matrix and an attribute matrix of nodes when it is computed by the encoder, as demonstrated in the Figure~\ref{figure:cell}. In practice, multiple identical neural cells are employed to build the entire network. Therefore, the search process is limited to a small cell, which necessitates balancing between efficiency and search cost.

\section{Implementation of DCLP}
\begin{figure}  
\centering  
\includegraphics[height=2.5cm,width=8cm]{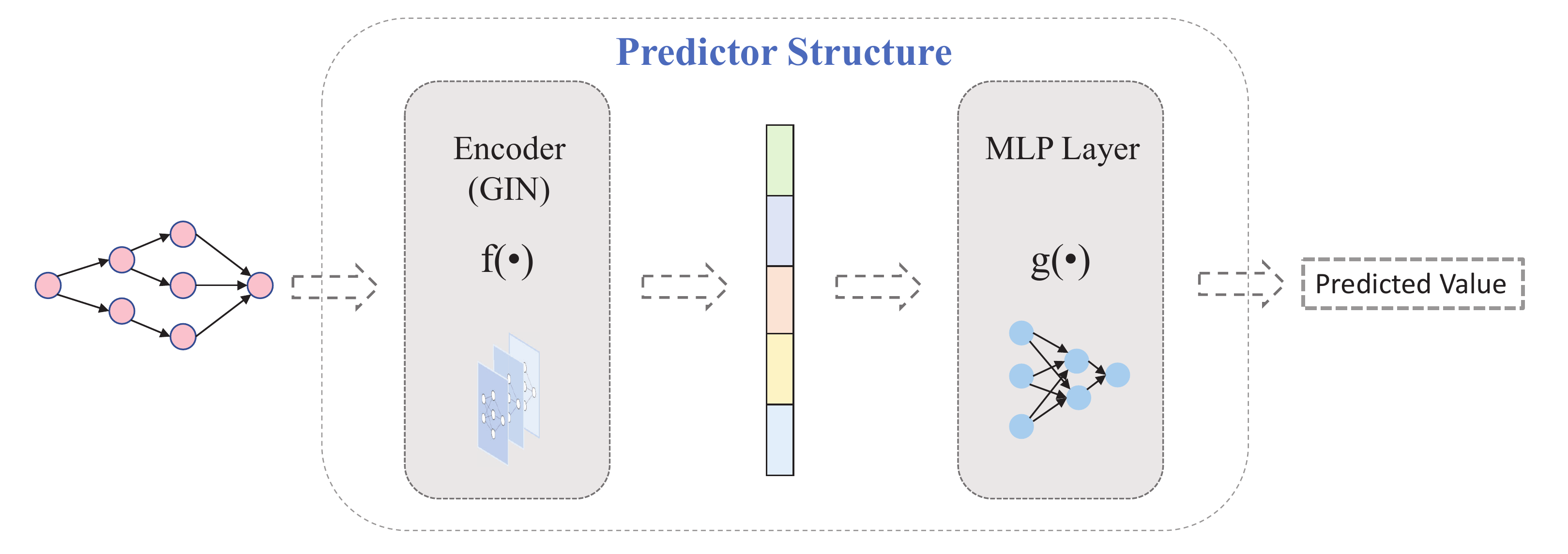}  
\caption{The structure of DCLP.}  
\label{figure:predictor}  
\end{figure}

In this section, we present the DCLP architecture, which consists of a GNN encoder and an MLP header. The predictor initially extracts features of the input neural architecture and represents it as a vector through the GNN encoder. To extract features from the DAGs, we use a three-layer GIN network $f(\cdot)$ since neural architectures are represented by DAGs. Each DAG is transformed into an N-dimensional feature representation after applying $f(\cdot)$. The MLP layer $g(\cdot)$ then transforms the N-dimensional vector into the predicted performance. The detailed structure diagram of DCLP is presented in Figure~\ref{figure:predictor}.

Note that the role of the MLP layer $g(\cdot)$ is limited to the prediction task. The pre-training using contrastive learning aims to train only the feature extraction ability, which involves learning the parameters of the encoder $f(\cdot)$. The encoder is a three-layer GIN network, which extracts features from the DAGs of neural architectures. For pre-training, assume that the positive data is $\left\{ q,k_+\right\}$ and the negative sample embeddings $\left \{ k_i\right\}_{i=1}^N$are saved in the memory bank. The contrastive learner's objective is to minimize the distance between $f(q)$ and $f(k_+)$, while maximizing the distance between $q$ and $k_i$. This learning method enables the encoder to learn encoding knowledge from unlabeled data. In fine-tuning, the complete predictor is formed by adding the MLP layer after the encoder.
\section{Hyper-parameter Settings}
We present the configuration details in two phases: pre-training and fine-tuning. During pre-training, we employ a batch size of 4096, 50 epochs, and a learning rate of 0.015. We set the dimension of the encoder output to 128, which is suitable for most neural cell sizes. Additionally, we use a stochastic gradient descent (SGD) optimizer to train the encoder based on the contrastive learning task.

We employ a three-layer MLP as $g(\cdot)$ in fine-tuning, with a curvature of 0.02 in LeakyReLU. We set the maximum epoch to 200 and employ early stopping based on loss. The batch size is adjusted dynamically based on the number of labeled training data; for instance, we use a batch size of 20 when there are 100 labeled data. To optimize the fine-tuning task, we use the Adam optimizer with a learning rate of 0.005. For other people's work, we utilize the default parameters and open-source code. We describe some DARTS search space-specific configurations in section E.2. We implement DCLP on an RTX 3090.
\section{Search Method}
In this section, we introduce the combination of DCLP with evolutionary (EA) and reinforcement learning (RL) algorithms to search for optimal architectures. The main approach involves sampling networks using a search strategy from the search space and utilizing DCLP to evaluate their performance, which guides further tuning of the search strategy. The significant advantage of using DCLP is the notable reduction in evaluation costs, leading to a reduction in the overall NAS process time. Further details can be found in the algorithm~\ref{alg:algorithm3}. In the main text, we have demonstrated the method combined with random search, while the algorithm presents the combination of two EA and RL techniques.
\begin{algorithm}[tb]
    \caption{Predictor-based Search Strategy}
    \label{alg:algorithm3}
    \begin{algorithmic}[1] %[1] enables line numbers
        \REQUIRE Search space $\mathcal{A}$, 
        fixed performance predictor $f$,
        the number of iterations $\mathcal{T}$,
        \STATE \textbf{Reinforcement Learning Strategy}
        \STATE Initialize t=0,an architecture set S=$\emptyset$
        \FOR{$t=1$ to $\mathcal{T}$}
        \STATE Sample $\mathcal{N}^t$ networks from  $\mathcal{A}$ using policy $\pi$,get  $S^t$.
        \STATE Evaluate each item $a \in S^t$ by trained predictor $f$
        \STATE Select top-k items from $\hat S^t$ according to $f(a)$, get $\widetilde S^t$
        \STATE $S \leftarrow S \cup \widetilde S^t$.
        \STATE Compute reward $R$,and update $\pi$.
        \ENDFOR
        \STATE \textbf{Evolution Strategy}
        \STATE Initialize t=0,an architecture set S=$\emptyset$ as population.
        \STATE Select $N_0$ architectures into S
        \FOR{$t=1$ to $\mathcal{T}$}
        \STATE Generate $\mathcal{N}^t$ architectures by mutating the architectures sampled from S,get $S^t$ .
        \STATE Evaluate each item $a \in S^t$ by trained predictor $f$
        \STATE Select top-k items from $\hat S^t$ according to $f(a)$, get $\widetilde S^t$
        \STATE $S \leftarrow S \cup \widetilde S^t$.
        \STATE Remove the old individuals from S
        \ENDFOR
        \STATE Evaluate each architecture in S by training to obtain the corresponding ground-truth $\mathcal{Y}$
        \ENSURE  the architecture $a^*$ with the best true performance.
    \end{algorithmic}
\end{algorithm}

\section{Discussion}
In this section, we discuss the validity of the approximation method employed in Section 3.2 to calculate the difficulty, as well as the rationale behind conducting normalization prior to utilizing MSE during the fine-tuning stage.
\subsection{The approximation of difficulty}
The rationality of the approximation in Section 3.2 can be attributed to the rationality of using the structural similarity approximation to measure the property similarity of neural networks. This is because we used WD to measure the difference between two uniform data augmentation distributions, which can be approximated by computing the integral of LD between each corresponding point on the two distributions. Here, each point on the distribution corresponds to a neural network, and LD measures the structural difference between the two neural networks. We will now discuss in detail the reasoning behind this approximation.

In our work, while acknowledging that the aforementioned approximation may have limitations, we find it to be a viable approach to achieve our objective. Specifically, this approximation is employed in pre-training to measure the complexity of a given positive data. Our objective is to map the neural network and its performance through the use of a predictor. To accomplish this goal, we employ an encoder to extract the characteristics of the neural architecture and encode them into an N-dimensional vector $z$, which represents the feature vector of the neural architecture. We then utilize an MLP to establish the relationship between $z$ and its corresponding performance. The fundamental principle guiding our approach is that as the gap between the performance values of two architectures decreases, so does the distance between their corresponding feature vectors.

\begin{figure}  
\centering  
\includegraphics[height=5.5cm,width=8.5cm]{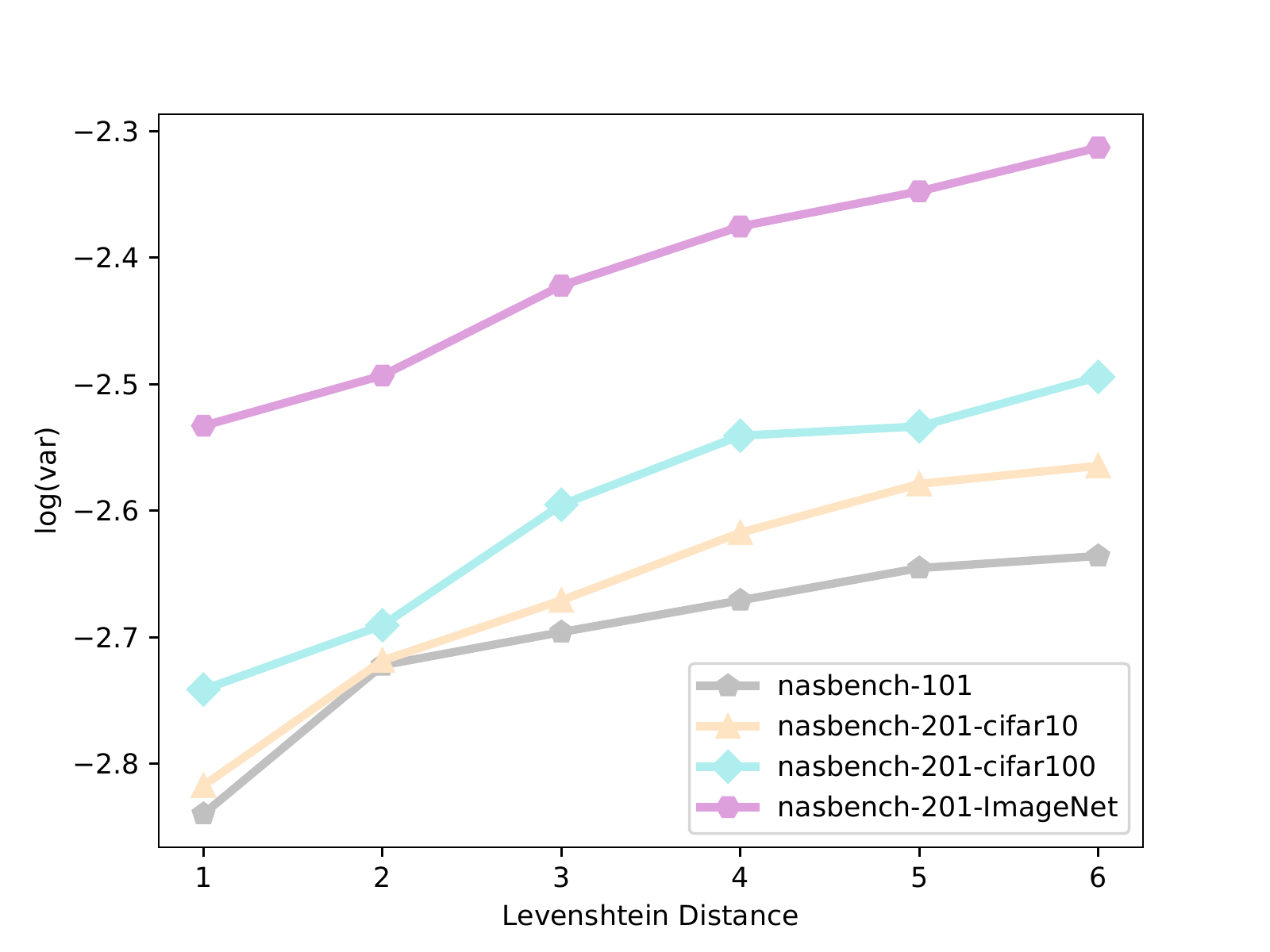}  
\caption{The variance of performance values of the architecture set with different LD on NASBench-101 and NASBench-201.}  
\label{figure:var}  
\end{figure}

In Figure~\ref{figure:var}, we demonstrate that as the LD between the data in the sampled neural network set increases, meaning that the structural difference is greater, the variance of the network's performance in the set also increases. This observation indicates that structural diversity can reflect the difference in the performance of neural architectures. In other words, the smaller the structural difference, the closer the performance of neural architectures, which is consistent with the characteristics of the neural network mentioned earlier. Thus, we can approximate the difference in features of the neural architectures by the difference in their structures.
\begin{figure}  
\centering  
\includegraphics[height=5.5cm,width=8.5cm]{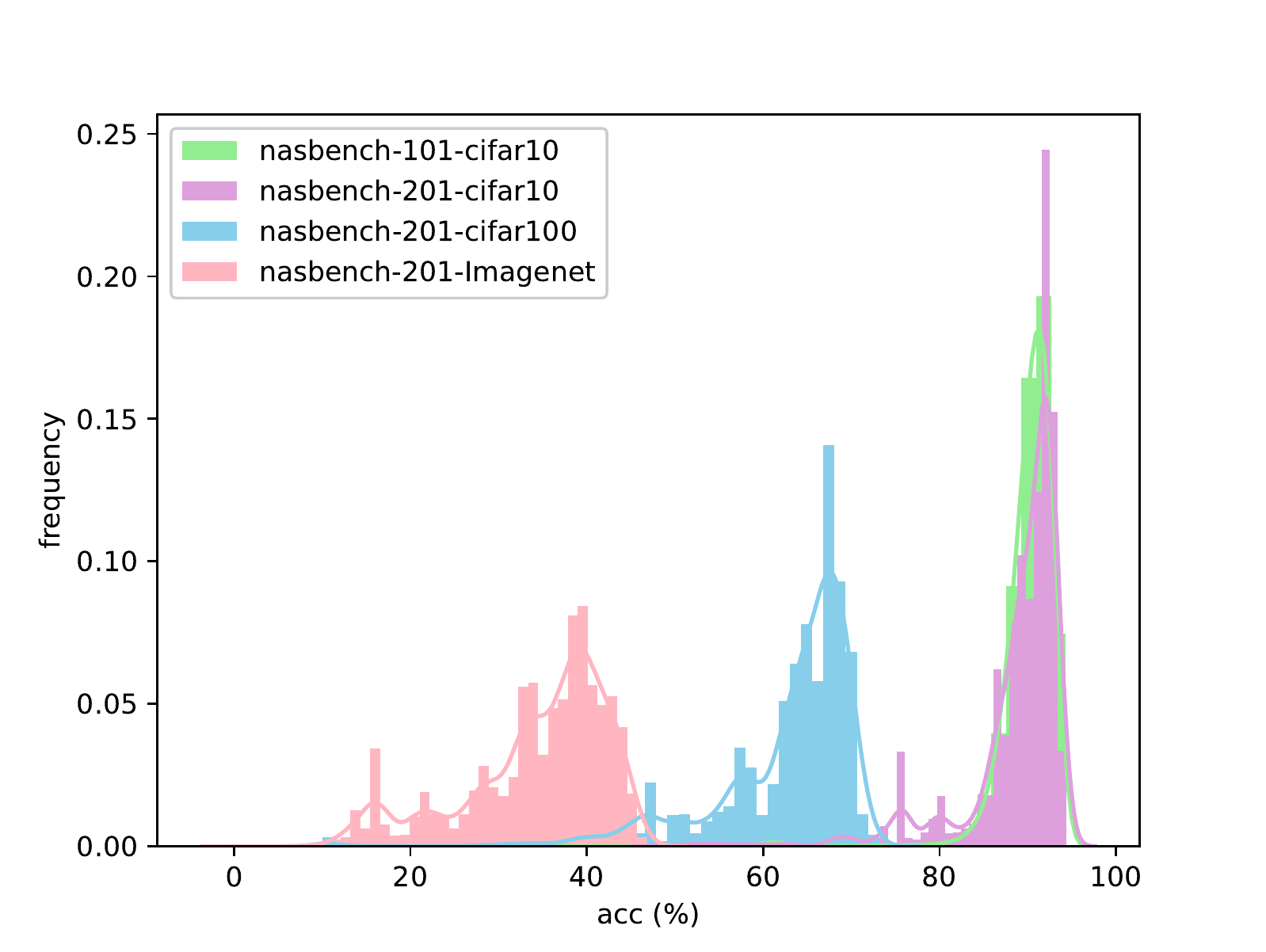}  
\caption{Distribution of architectures’ performance values on NAS-Bench-101 and NAS-Bench-201.}  
\label{figure:performance}  
\end{figure} 
\subsection{Why normalization is useful for MSE}
In the main text, we have briefly discussed the optimal effect of normalization on MSE in our task. This section aims to delve deeper. We argue that MSE is not suitable for our fine-tuning process since a tiny MSE value may not always correspond to high sorting accuracy, which is our main concern. Consider two architectures in the training set, where the performance of each neural architecture is a float number between 0 and 1. If the performance difference between the two networks is only 0.01, and the predictor evaluates the performance of the two architectures as precisely the opposite, the ranking of the two networks changes. However, the MSE in this case is still small. Therefore, the optimization goal of vanilla MSE is not consistent with our objective to maximize the ranking accuracy.

Note that the performance distribution of neural networks in the search space, as shown in Figure~\ref{figure:performance}, is approximately normal. This means that there are few architectures with extremely good or poor performance, and most networks perform within a certain range. Therefore, the performance gap between many architectures is small, making it difficult to learn ranking information using vanilla MSE. Therefore, we use normalization to enhance MSE, such as by using the Z-score. It maps the original normal distribution, ranging between 0 and 1, to a standard normal distribution. The formula used is as follows:
\begin{equation}
x^*=\frac{x-\overline{x}}{\sigma}
\end{equation}
where $\overline{x}$ is the mean of the original performance, and $\sigma$ is the standard deviation. When there are two architectures with original performances $x_i$ and $x_j$, and the performance difference is $\delta$ which is a small value, it is difficult to learn the relative ranking relationship of these two architectures using MSE directly. However, after normalization, the performance difference between the normalized values of $x_i$ and $x_j$ becomes $\frac{\delta}{\sigma}$. As $\sigma$ is a positive number less than 1, the distance between two similar performances is artificially increased. Thus, if the predictor predicts a deviation in the ranking of two networks with similar original performance, there is a large gap between the normalized performance, so the MSE value is also large. This helps to unify the optimization goal of MSE with the target for ranking accuracy. Therefore, in our task, using MSE+normalization is adequate for enhancing the ranking accuracy.
\begin{figure}
	\centering
        \subfigbottomskip=-2 pt 
	\subfigcapskip=-5pt 
	\subfigure[SSANA]{
		\begin{minipage}[t]{0.5\linewidth}
			\centering
			\includegraphics[width=1\textwidth]{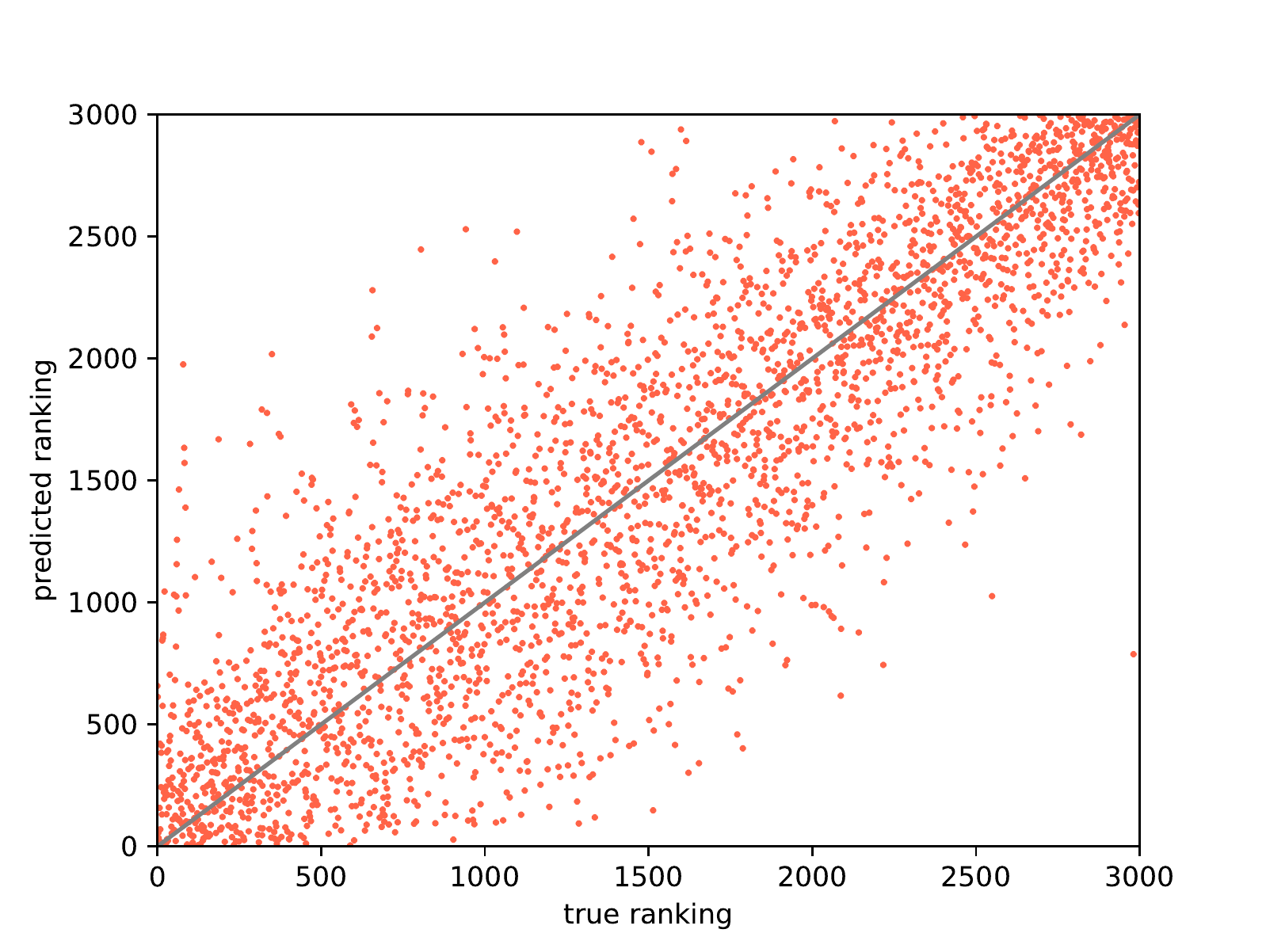}
		\end{minipage}
	}%
	\subfigure[HAAP]{
		\begin{minipage}[t]{0.5\linewidth}
			\centering
			\includegraphics[width=1\textwidth]{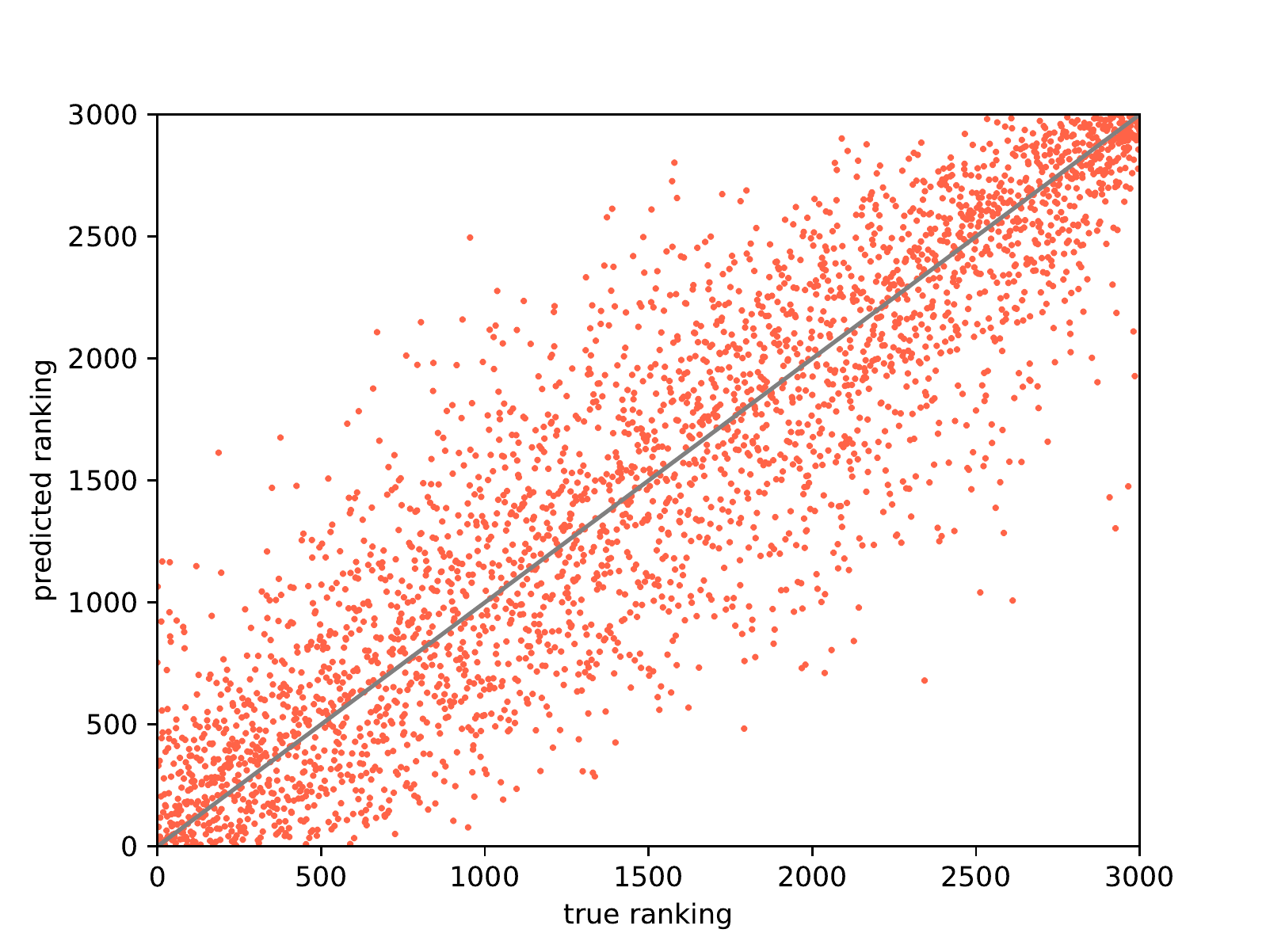}
		\end{minipage}
	}%
	\\
	\subfigure[GMAE]{
		\begin{minipage}[t]{0.5\linewidth}
			\centering
			\includegraphics[width=1\textwidth]{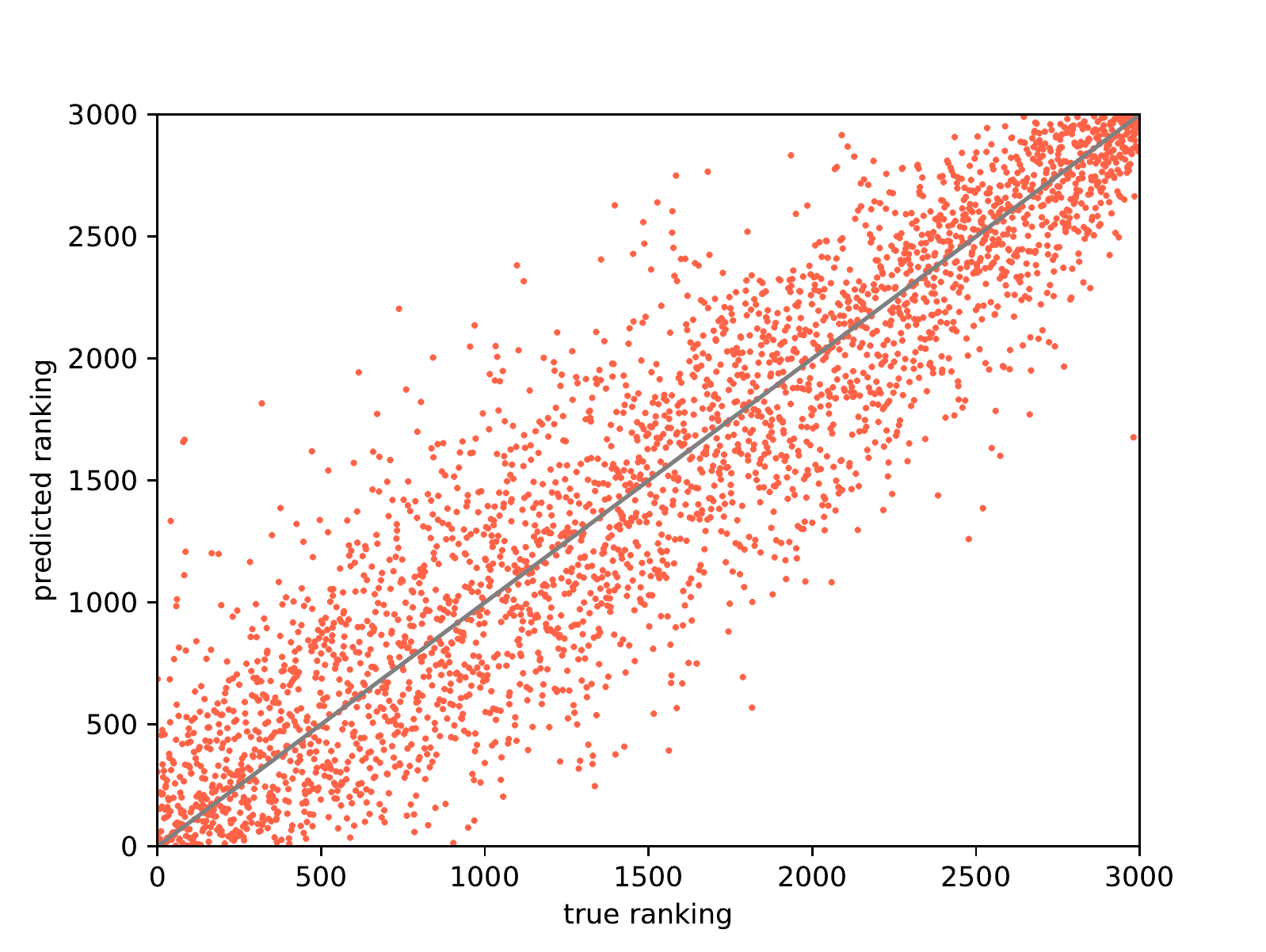}
		\end{minipage}
	}%
	\subfigure[DCLP]{
		\begin{minipage}[t]{0.5\linewidth}
			\centering
			\includegraphics[width=1\textwidth]{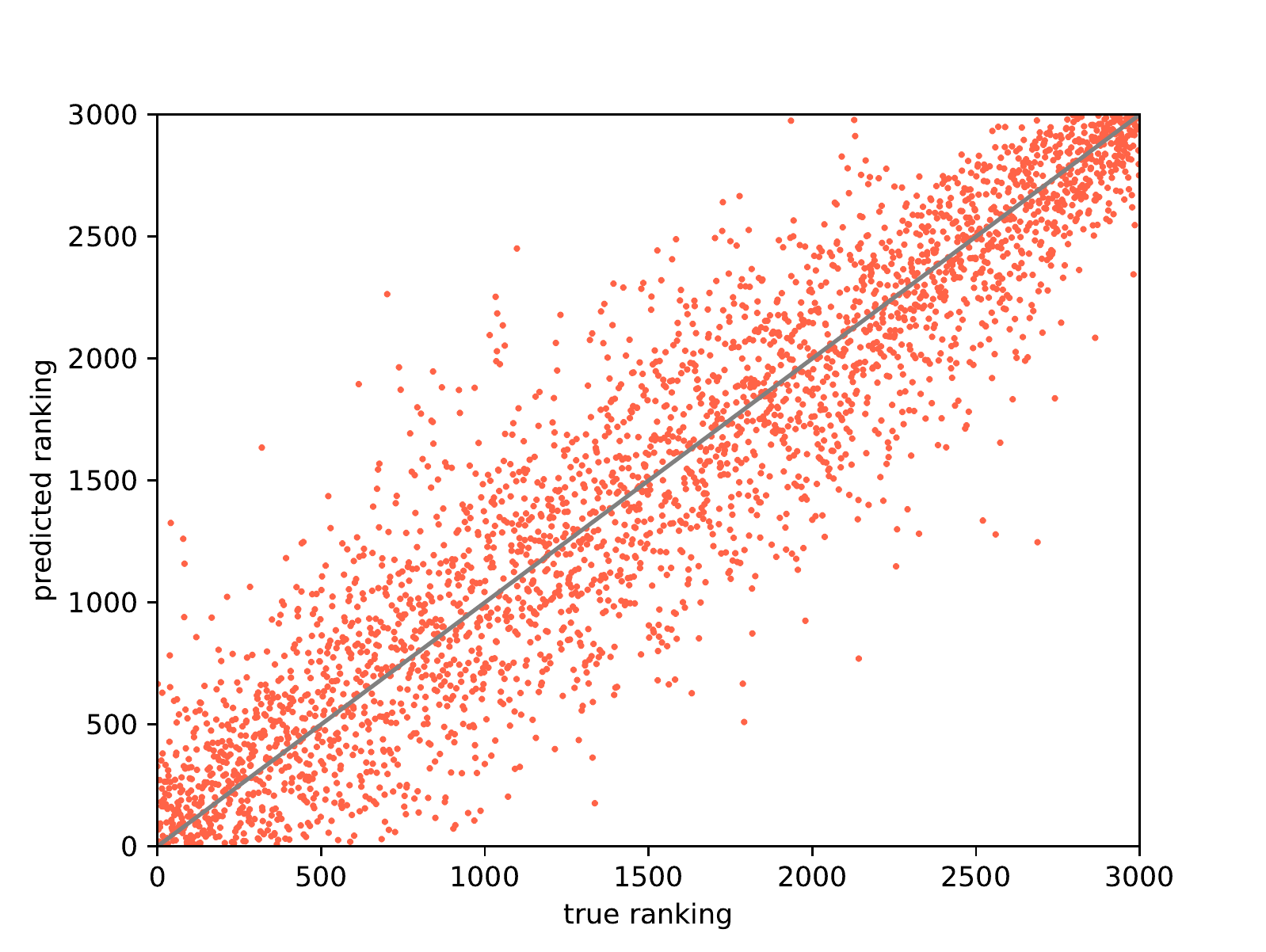}
		\end{minipage}
	}%
	\centering
	\caption{Comparison with state-of-the-art methods on NAS-Bench-101. 300 trained architectures are used to train the neural predictors.}
	\label{ktau}
\end{figure}

\section{Comparison with CTNAS}
\begin{table}[htb]
  \centering
    \begin{tabular}{ccc}
    \toprule
    parameter & Search & Evaluation \\
    \midrule
     batch size & 96 & 96\\
     layers & 8 & 20\\
     learning rate & 0.025 & 0.025\\
     epochs & 50 & 600\\
     init channels & 32 & 36\\
     momentum & 0.9 & 0.9\\
     weight decay& 0.0003 & 0.0003\\
     auxiliary weight& 0.4 & 0.4\\
     drop path prob & 0.2 & 0.2\\
     cutout length & 16 &16 \\
     grad clip & 5 & 5\\
    \bottomrule
    \end{tabular}%
  \caption{The DARTS setups during search and evaluation in image classification task.}
  \label{tab:darts}%
\end{table}%
As previously discussed, the utilization of contrastive learning and curriculum learning in CTNAS may appear to be similar to our approach. However, there are fundamental differences between the two methods. You could even say there is no connection. The subsequent section elaborates on the dissimilarities in both contrastive learning and curriculum learning techniques.

DCLP utilizes contrastive learning as an unsupervised technique that leverages similar and dissimilar samples for pre-training. In contrast, CTNAS requires a neural performance comparator that can predict which of the two neural networks performs better and uses labeled data for training, and this prediction is referred to as contrastive. Therefore, it differs significantly from DCLP in both motivation and implementation.

The use of curriculum learning in CTNAS is distinct from its typical meaning. Unlike traditional curriculum learning, which schedules training data to enhance training efficiency, CTNAS uses curriculum to update the NAS search results by choosing the best neural network. Conversely, DCLP employs curriculum learning to schedule training data for stable pre-training. Thus, both the starting point and implementation of the curriculum in CTNAS and DCLP differ.

In conclusion, CTNAS and DCLP originate from two distinct starting points in the domain of neural predictors. So our work is innovative, not an imitation of CTNAS.
\section{Additional Experimental Results}
In this section, we mainly show some visualization results of the performance of DCLP on NAS-Bench-101 and the optimal architectures searched on NAS-Bench-101 and DARTS search spaces.
\subsection{Results on NAS-Bench-101}
The qualitative comparison of different neural predictors with Kendall's Tau is shown in Figure~\ref{ktau}. The test set consists of 3,000 randomly sampled structures, where the $x$-axis represents the actual ranking and the $y$-axis represents the predicted ranking of the architecture. The correlation between the rankings can be observed by the proximity of the points to the diagonal, with higher Kendall's Tau corresponding to a stronger ranking correlation. Our approach shows better results with 300 training data, as observed in the figure. This visualization allows for a more intuitive understanding of the performance of predictors compared to using the values of Kendall's Tau alone.
\begin{table}[htbp]
  \centering
  \scalebox{0.75}{
    \begin{tabular}{ccccc}
    \Xhline{1pt}
    \multirow{2}{*}{Method} & \multirow{2}{*}{Top1(\%)} & \multirow{2}{*}{\makecell[c]{Cost\\(GPU h)}}& \multirow{2}{*}{Type}\\
    & & & \\
    \Xhline{0.4pt}
    RFGIAug~\cite{xie2022architecture}& 73.4&  36 & pred+RS\\
    CATE~\cite{yan2021cate}& 73.9& 75& pred+RS\\
    RANK-NOSH~\cite{wang2021rank} & 74.8& 48.2& RANK-NOSH\\
    $\beta$-DARTS~\cite{ye2022b}& 75.8&9.6&supernet\\
    DARTS-~\cite{chu2020darts}& 76.2&108&supernet\\
    \Xhline{0.4pt}
    DCLP+RS& 76.9& 6& pred+RS\\
    \Xhline{1pt}
    \end{tabular}%
    }
    \caption{The comparison of NAS in DARTS space with ImageNet as the dataset. \textit{pred} means predictor.}
  \label{tab:DARTS}%
\end{table}
\begin{table}[htbp]
  \centering
    \begin{tabular}{ccc}
    \toprule
    parameter & Search & Evaluation \\
    \midrule
     batch size & 256 & 64\\
     layers & 12 & 12\\
     learning rate & 20 & 20\\
     epochs & 50 & 600\\
     embedding size & 300 & 850\\
     clip& 0.25 & 0.25\\
     sequence length& 35 & 35\\
     weight decay & 5e-7 & 8e-7\\
     dropout & 0.75 &0.75 \\
    \bottomrule
    \end{tabular}%
  \caption{The DARTS setups during search and evaluation in the language model task.}
  \label{tab:darts_RNN}%
\end{table}%

\begin{figure*}  
\centering  
\includegraphics[height=8cm,width=16cm]{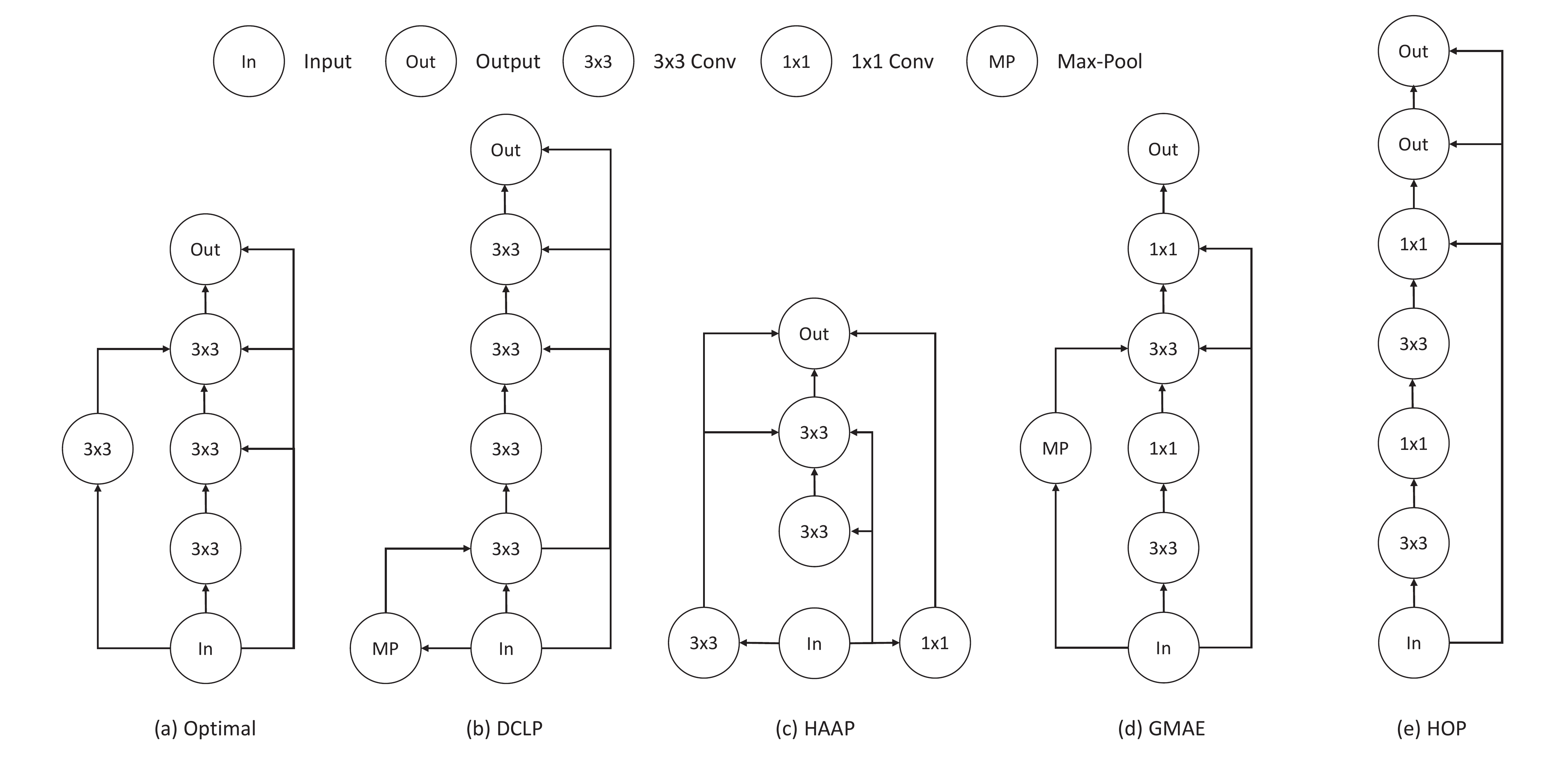}  
\caption{Visualization of the best network architectures selected by different methods. 300 architectures randomly selected from NAS-Bench-101 are used as labeled examples.}  
\label{figure:NAS-101-arch}  
\end{figure*}

Figure~\ref{figure:NAS-101-arch} shows the visualization of the structures searched on NAS-Bench-101 using 300 labeled training data by different predictor-based methods.
\subsection{Results on DARTS search space}
In this section, we discuss the parameter settings and search results on the darts search space in two tasks: image classification and language model.

\begin{figure}  
\centering  
\includegraphics[height=4.5cm,width=8.5cm]{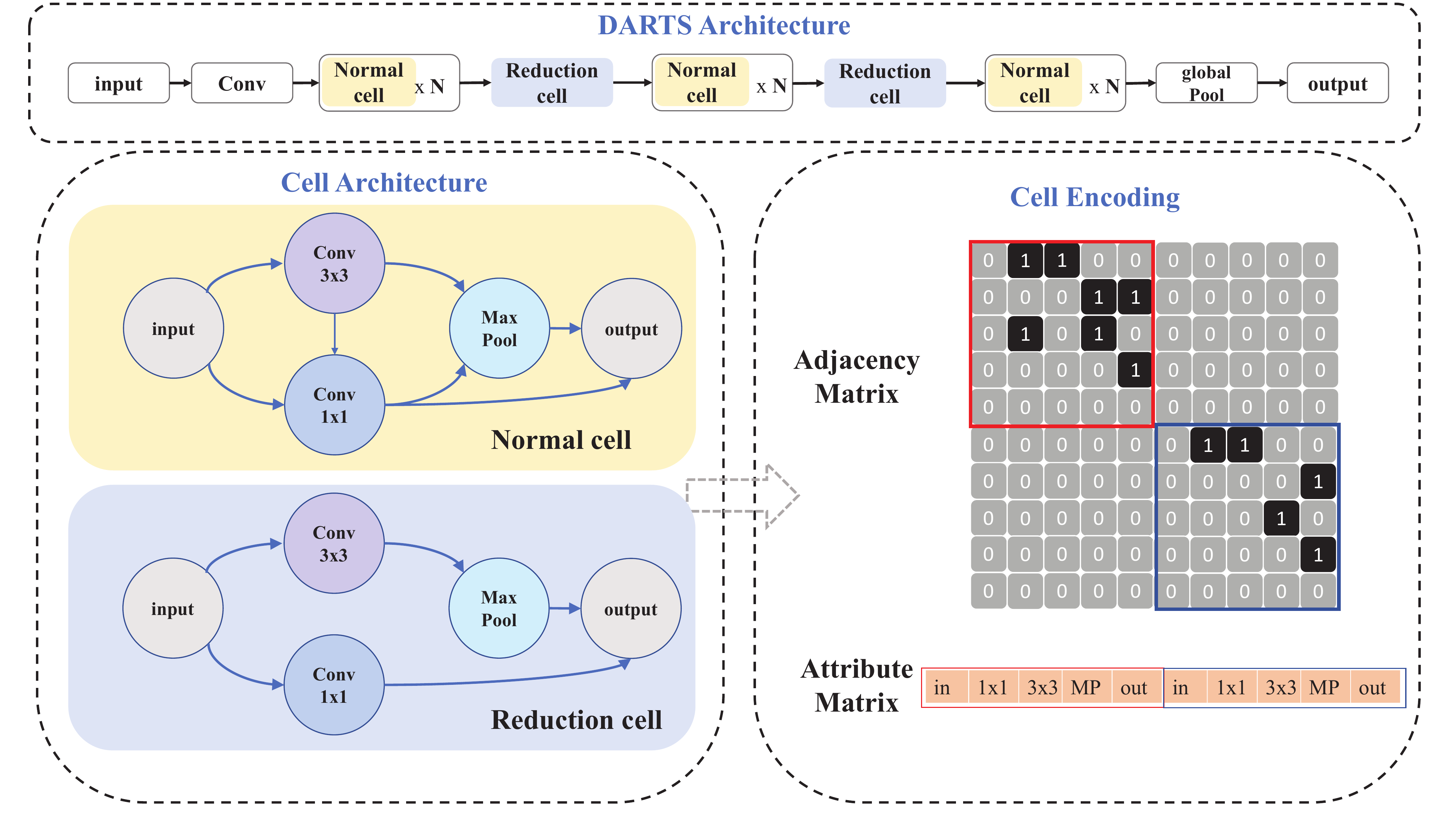}  
\caption{Schematic diagram of the cell structure when searching in DARTS space. The normal cell and the reduction cell are combined into a single cell for searching.}  
\label{figure:search_space}  
\end{figure}

\begin{figure*}
	\centering
        \subfigbottomskip=-2 pt 
	\subfigcapskip=-5pt 
	\subfigure[Normal Cell]{
		\begin{minipage}[t]{0.5\linewidth}
			\centering
			\includegraphics[width=1\textwidth]{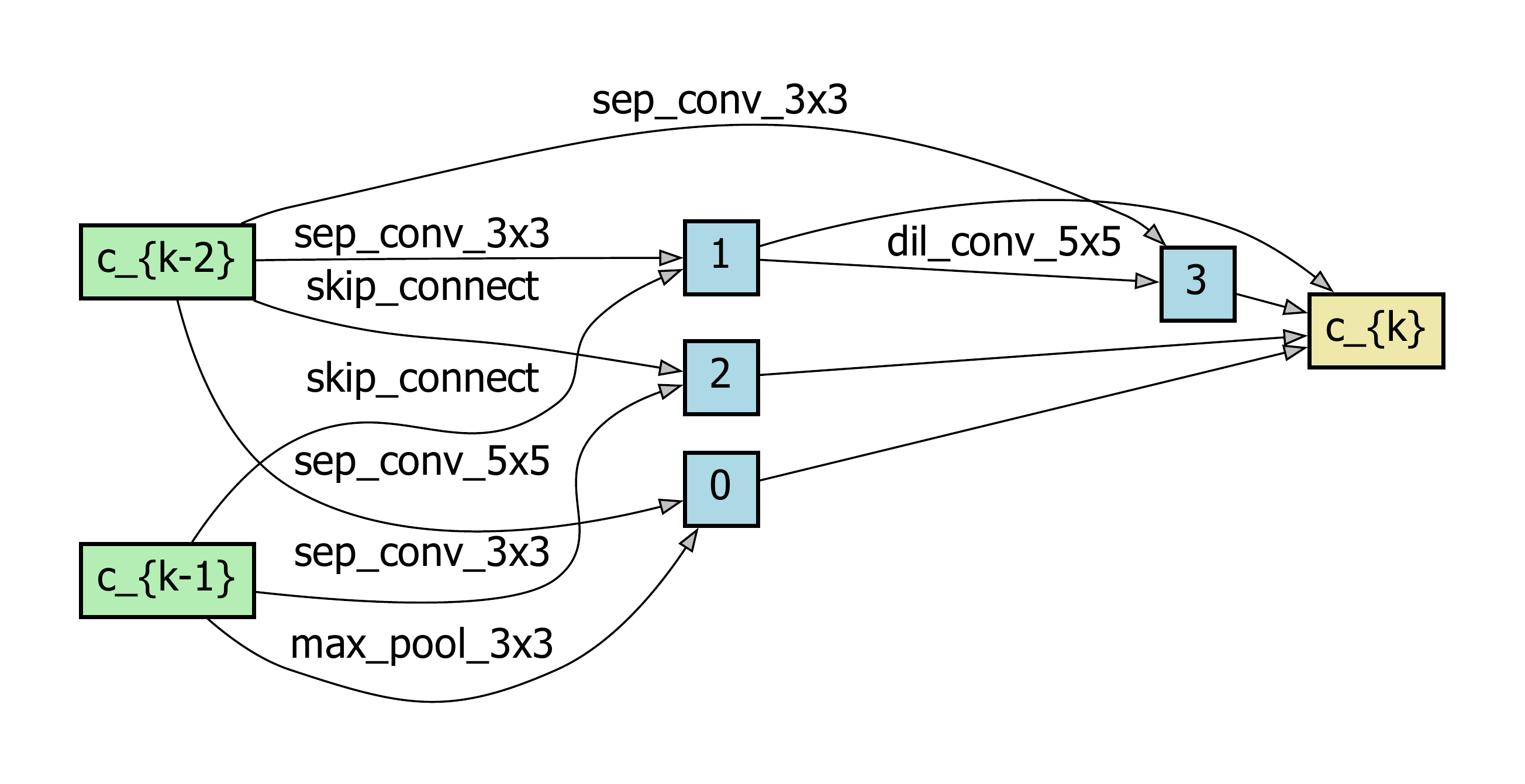}
		\end{minipage}
	}%
	\subfigure[Reduction Cell]{
		\begin{minipage}[t]{0.5\linewidth}
			\centering
			\includegraphics[width=1\textwidth]{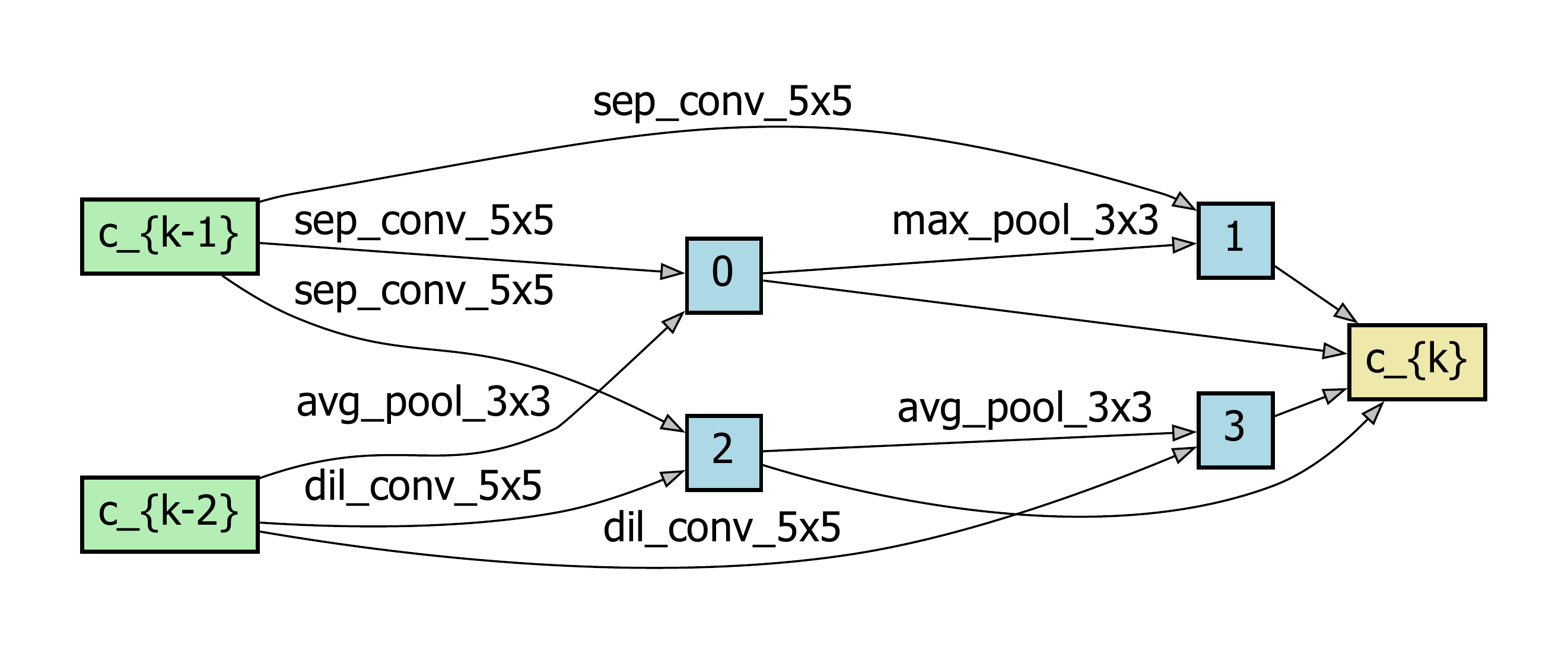}
		\end{minipage}
	}%
	\centering
	\caption{The best-discovered architecture of the image classification task on DARTS search space.}
	\label{darts}
\end{figure*}

\textbf{Image classification.} In contrast to NAS-Bench-101, the DARTS search space in the image classification task involves searching for two types of cells, namely normal and reduction cells. In order to obtain the performance of both cells in a single iteration, we combine them into a single cell for searching. A schematic diagram of this process is presented in Figure~\ref{figure:search_space}, while Figure~\ref{darts} illustrates the optimal structure that was discovered on the DARTS search space through the combined use of DCLP and random search methods on the CIFAR10 dataset. For fairness in comparing performance with other works, we followed the DARTS hyperparameter setup~\cite{liu2018darts}, which is shown in Table~\ref{tab:darts}. It is important to note that during the acquisition of training data for DCLP, we only required a rough estimation of the performance ranking relationship between architectures, so to minimize training overheads, we only used the results from the first three to five epochs of training as the performance metric for the architectures in the training set.

\begin{table}[htbp]
  \centering
  \scalebox{0.9}{
    \begin{tabular}{cccc}
    \Xhline{1pt}
    Method & Acc(\%) & Ranking(\%) &Query\\
    \Xhline{0.4pt}
    embedding similarity&93.65±0.18&0.28 &300\\
     WD (Ours)&94.17±0.06&0.0014& 300\\
     \Xhline{1pt}
    \end{tabular}%
    }
    \vspace{-0.2em}
  \caption{The ablation result of similarity measurements.}
  \label{tab:difficult measurement}%
\end{table}
\begin{table}[htb]
  \centering
  \scalebox{1}{
    \begin{tabular}{cccc}
    \Xhline{1pt}
    Method & Acc(\%) & Ranking(\%) &Query\\
    \Xhline{0.4pt}
    GCN &93.98±0.05& 0.015 & 300\\
    GAT &93.79±0.08& 0.12 & 300\\
    GIN (Ours)&94.17±0.06&0.0014& 300\\
     \Xhline{1pt}
    \end{tabular}%
    }
    \vspace{-0.2em}
  \caption{The ablation result of graph representations.}
  \label{tab:graph representation}%
\end{table}

Also, we search on ImageNet using the Darts search space. The experimental results of the search are shown in table~\ref{tab:DARTS}. As the data shows, DCLP also shows good capability on ImageNet. This proves that DCLP is not only suitable for small datasets like CIFAR-10 but also for larger datasets like ImageNet. This group of trials provides even more compelling evidence for the superiority of our methods. 

\begin{figure*}  
\centering  
\includegraphics[width=0.5\linewidth]{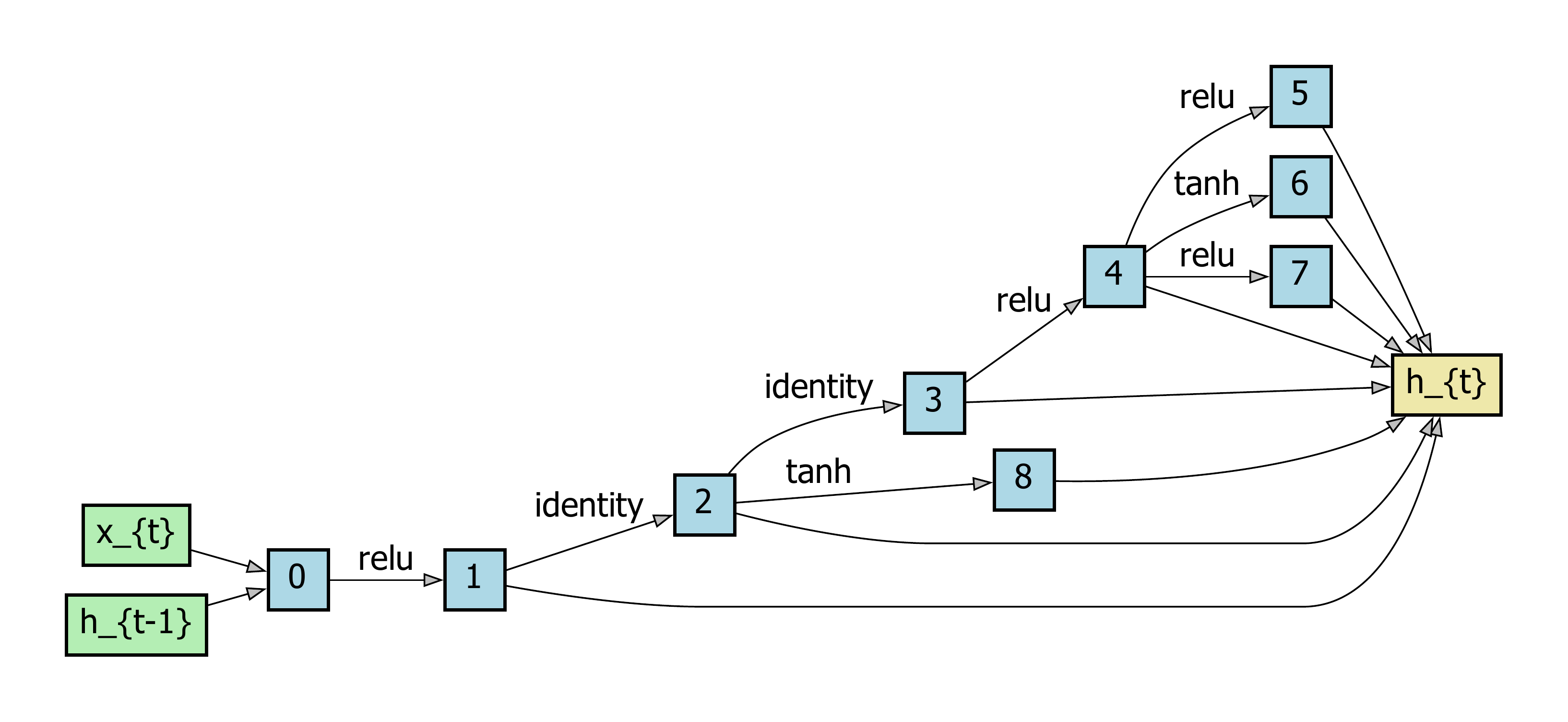}  
\caption{The best-discovered architecture of the language model task on DARTS search space.}  
\label{figure:darts_rnn}  
\end{figure*}

\textbf{Language model.}
In the language model task, the search space consists of an RNN network where only one cell is searched during the search. The parameters are set up the same as DARTS~\cite{liu2018darts} and can be viewed in the table~\ref{tab:darts_RNN}. The task is carried out on the PTB dataset, and similar to the previous experiments, we use the training results of five epochs as the training set to train DCLP. Figure~\ref{figure:darts_rnn} displays the search results obtained using DCLP combined with random search. We experiment in the space of both CNNs and RNNs and demonstrate the generalizability of DCLP across multiple neural networks.

\subsection{Ablation experiment}
To evaluate different measurements, we conducted experiments on NAS-Bench-101 as table~\ref{tab:difficult measurement}. The embedding similarity uses GNN as the encoder and utilizes the embedding similarity as the neural network similarity. We find that structural similarity surpasses embedding similarity, indicating its superiority in measuring neural network similarity.

All GNNs can serve as graph representations. We conduct a comparison of GCN, GAT, and GIN on NAS-Bench-101 as Table~\ref{tab:graph representation}. Among these models, GIN excels in performance due to its strong ability to distinguish graph isomorphisms, a critical factor highlighted in the paper.

\end{document}